\documentclass[sci2]{roampreprint}
\runningauthor{Miki et al.}
\kicker{Research article}
\summary{Vision-language models give legged robots the common sense to hike kilometers autonomously in alpine and urban terrain.}
\links{Project page: \url{https://roam-hiking.github.io} \quad Movie 1: \url{https://youtu.be/xcqVpbgklD4} \quad Field documentary: \url{https://youtu.be/wrbAq2HXE8E}}

\usepackage{amsmath}
\usepackage{amssymb}
\usepackage{multirow}
\usepackage{booktabs}
\usepackage{float}
\usepackage{xcolor}
\usepackage{url}
\usepackage{bm}
\usepackage{listings}
\usepackage{longtable}
\usepackage{siunitx}
\usepackage{verbatim}
\usepackage[nohyperlinks, printonlyused, nolist]{acronym}

\setboolean{shortarticle}{false}

\DeclareMathOperator*{\argmax}{arg\,max}
\newcommand{\states}{\mathcal{X}}
\newcommand{\Measurements}{\mathcal{Z}}

\newcommand{\scititle}{Bridging vision-language reasoning and learned control for legged navigation in the wild}
\title{\scititle}

\author[1]{Takahiro Miki}
\author[1]{Robin Schmid}
\author[1]{Manthan Patel}
\author[1]{Turcan Tuna}
\author[1]{Julia Richter}
\author[1]{Julian Nubert}
\author[1]{Joonho Lee}
\author[2]{Matthias M\"uller}
\author[1]{Marco Hutter}

\affil[1]{ETH Zurich, Zurich, Switzerland}
\affil[2]{Apple, Cupertino, CA, USA}

\dates{}
\begin{abstract}
Autonomous navigation in diverse real-world environments remains one of the greatest challenges in robotics. It demands both robust low-level motion control to handle diverse terrain and high-level reasoning to interpret environmental cues. We present ROAM (Robust Open-world Autonomous Mobility), a system that enables legged robots to perform kilometer-scale navigation in complex alpine and urban settings. ROAM combines the world knowledge of a Vision-Language Model (VLM), pre-trained on internet-scale data, with robust locomotion and local planning policies learned in simulation via reinforcement learning (RL). The VLM provides common-sense reasoning and semantic interpretation of the scene, assessing risks, interpreting signs, and producing semantic labels with associated navigation costs. An open-vocabulary segmentation module converts these labels into a dense, real-time risk map, which conditions two pre-trained RL policies responsible for agile locomotion and local navigation, effectively coupling high-level VLM reasoning with low-level physical control. Real-world experiments showed the system's ability to autonomously navigate a wide range of complex real-world environments. The robot completes kilometer-long hikes with minimal intervention, interprets signs to avoid danger, and evaluates its environment to identify safe areas on which to step. This work represents a step toward the reliable deployment of systems in unstructured environments. However, it also underscores the challenges that remain in VLM reliability and spatial reasoning, indicating that foundational models are a crucial component but not yet a complete solution for navigation in the wild.
\end{abstract}

\begin{document}

\begin{acronym}
\acro{AI}[AI]{Artificial Intelligence}
\acro{LLM}[LLM]{Large Language Model}
\acro{VLA}{vision-language-action}
\acro{VLM}{Vision-Language Model}
\acro{RT}{Real-Time}
\acro{GT}{Ground Truth}
\acro{GNSS}{Global Navigation Satellite System}
\acro{RTK}{Real-time Kinematic Positioning}
\acro{LiDAR}{Light Detection and Ranging}
\acro{IMU}{Inertial Measurement Unit}
\acro{RPE}{Relative Position Error}
\acro{RRE}{Relative Rotation Error}
\acro{APE}{Absolute Position Error}
\acro{ARE}{Absolute Rotation Error}
\acro{RL}{Reinforcement Learning}
\acro{HF}{Holistic Fusion}
\acro{MAP}{Maximum a Posteriori}
\acro{SLAM}{Simultaneous Localization and Mapping}
\end{acronym}

\maketitle

\section{Introduction}
Enabling robots to navigate autonomously and safely across diverse, unstructured environments --- from indoor spaces and urban areas to complex natural terrain --- is a fundamental challenge in robotics. Although recent advances in legged robotics have demonstrated robust locomotion in the wild~\cite{raibert2008bigdog,lee2020learning,miki2022wild,Tranzatto2022-sm,choi2023learning,lee2024learning,miki2024confined,unitreerobotics2024,deeprobotics2024,limxdynamics2024}, the ability to autonomously navigate these environments remains an unresolved challenge.

Traditional navigation methods, which rely heavily on geometric traversability estimation, often fail in the presence of non-geometric hazards such as tall grass, water, or mud. Furthermore, these systems disregard implicit  semantic norms, such as staying on a designated trail, which are crucial for safe and acceptable navigation in environments shared with people. Although semantic segmentation-based approaches aim to provide high-level understanding, their reliance on fixed datasets or label sets restricts their adaptability across the diverse and varied conditions encountered in the real world.

 A compelling hint to what these autonomous systems are missing can be found by observing a human remotely operating the same robot. Even without prior experience, a person can successfully guide the robot through complex environments. This indicates that the bottleneck is not the robot's ability to move, but its ability to decide where to move. Humans accomplish this by applying common-sense reasoning: we instinctively understand that a paved path is safe, that deep puddles are a risk, and that one should follow a trail. This intuitive, real-world understanding is the crucial component that current navigation systems lack.

Recent breakthroughs in generative AI offer a promising way to equip robots with this missing common-sense reasoning. Foundation models like \acp{LLM}, trained on internet-scale data, have shown a strong capacity for high-level reasoning~\cite{radford2018improving,devlin2018bert,brown2020language,touvron2023llama,liu2024deepseek,yang2025qwen3}. \acp{VLM} have extended this capability into the visual domain~\cite{team2023gemini,bai2023qwen,lu2024deepseek}. Meanwhile, self-supervised visual foundation models have enabled powerful image understanding~\cite{radford2021learning,oquab2023dinov2,ravi2024sam}.
 By learning from vast quantities of image and text data, these models can interpret visual scenes, understand concepts such as trail or puddle, and reason about their implications for navigation. This ability to perform common-sense reasoning directly from visual input makes \acp{VLM} a compelling candidate for overcoming the limitations of traditional navigation systems.

This paper investigates how the common-sense reasoning inherent in \acp{VLM} can be utilized for autonomous navigation in the wild. We propose and evaluate a framework that translates the high-level understanding of these models into robust navigational policies, thereby assessing the readiness of this approach for deployment on real-world robotic systems.

\subsection{\acp{VLM} for robot navigation}
The key challenge in applying \acp{VLM} to robotics is bridging their high-level, common-sense reasoning with the low-level commands required for physical control. An embodied agent must ultimately translate multimodal sensory inputs, such as images and proprioception, into joint-level commands (such as target positions or torques) to produce movement.

The most direct approach is to have the \ac{VLM} generate joint-level actions. This can be attempted either by directly prompting the model or by fine-tuning the pretrained \ac{VLM} into a \ac{VLA} model, which adapts the architecture with specialized action heads trained on embodied datasets. 
This has been explored for tasks like manipulation~\cite{rt12022arxiv,brohan2023rt,open_x_embodiment_rt_x_2023,kim2024openvla}. 
These end-to-end models directly control the robot's actuators from image inputs. However, this approach faces hurdles: it typically necessitates massive datasets of human demonstrations, which are costly to acquire for diverse environments and difficult for specific robot embodiments such as quadrupeds. Additionally, the high-frequency control required for dynamic tasks, such as locomotion, presents a challenge for the onboard deployment of large models.

To improve data efficiency and reactivity, a common strategy is to raise the level of abstraction for the robot's action space, making the task easier for the \ac{VLM} and improving generalization across different robot embodiments. These abstract commands can then be converted into joint-level actions through traditional controllers or other learning-based modules, such as reinforcement learning policies.
A central consideration is identifying the level of abstraction that best supports \ac{VLM}-driven navigation across diverse environments.
One level of abstraction involves having the \ac{VLM} or foundation model directly generate continuous waypoints~\cite{shah2023vint,sridhar2024nomad,hirose2024lelan,zhou2025autovla}, or trajectories~\cite{gu2023rt, windecker2025navitrace}. Although this leverages the model's reasoning, it requires consistently interpreting spatial information from images and generating feasible actions from a high-dimensional space. For example, projecting waypoints in the image space to the Euclidean command space requires depth estimation and geometric post-processing.
Transforming VLM outputs into robot commands remains a challenge for current approaches~\cite{windecker2025navitrace, ramakrishnan25}.

To further simplify the task, another line of research uses the \ac{VLM} as a high-level planner to select from a discrete set of candidate actions. These candidates can be labeled target positions~\cite{nasiriany2024pivot,goetting2024end,sathyamoorthy2024convoi,elnoor2024robot}, potential trajectories~\cite{song2024tgs,liang2025mosu}, or abstract instructions like landmarks~\cite{zhou2024navgpt,zhang2024navid,cheng2024navila,weerakoon2024behav}. By narrowing the action space, the \ac{VLM} only needs to identify the best option, rather than generate a precise command. However, these methods often introduce high latency, as each step requires querying the \ac{VLM}. Although some works incorporate spatial maps for indoor tasks~\cite{yokoyama2024vlfm,huang2023visual,wei2024ovexp,chiang2024mobility}, applying this paradigm to large, complex outdoor settings remains an open challenge.

These VLM-centric strategies stand in contrast to traditional outdoor navigation systems. For decades, these systems have employed traversability estimation methods that use geometric information~\cite{thrun2006stanley,fieldlocalplanner2022,Fankhauser2016GridMapLibrary, Fanstep}, or leverage models trained on data from human annotations~\cite{wellhausen2021rough,schreiber2024w}, predefined heuristics~\cite{freyroadrunner, patel2024roadrunner}, or simulated data tailored to specific robot embodiments~\cite{yang2021real,frey2022locomotion}. 

 These approaches usually estimate traversability in 2.5D elevation maps consumed by planners such as model predictive path integral (MPPI) control~\cite{williams2017model} for navigation. Expanding on these geometric methods, recent work has also tackled discrete terrain navigation, where systems must identify highly precise footholds to traverse disjointed surfaces, gaps, or stepping stones~\cite{kim2025high, he2025attention}.

Additionally, self-supervised learning techniques have been explored, using various forms of supervision such as traction estimates~\cite{gasparino2022wayfast, frey23fast, gasparino2024wayfaster}, collision feedback and bumpiness~\cite{kahn2021badgr}, trajectory-based guidance~\cite{wellhausen2019should,schmidtrav, jung2024v}, and human navigation demonstrations~\cite{kim2024learning}. Although these approaches often build spatial representations for real-time planning---thereby avoiding the latency of continually querying large language models---they lack the contextual adaptability that common-sense reasoning provides. They tend to depend on fixed datasets, struggle to generalize to unseen terrains, and require retraining whenever the robot platform or operating conditions change.

This presents a trade-off: VLM-based methods offer rich semantic understanding but face challenges in latency, data requirements, or grounding for direct control, whereas traditional methods are efficient but semantically brittle. This motivates a hybrid approach that seeks to combine the strengths of both paradigms. Although recent work has taken steps in this direction~\cite{weerakoon2024behav}, its reliance on scene-specific user input restricts its use in large-scale autonomous navigation, since the LLM prompt specifying where to stay and where to avoid must be manually provided by the user for every new scene.

In this paper, we introduce a system for robust open-world autonomous mobility (ROAM), a hybrid approach to investigate the extent to which a \ac{VLM}'s common-sense reasoning can enable autonomous navigation in the wild (\href{https://youtu.be/xcqVpbgklD4}{Movie 1}).

\subsection{The ROAM system}
\begin{figure*}[tbp] %
    \centering
    \includegraphics[width=\textwidth]{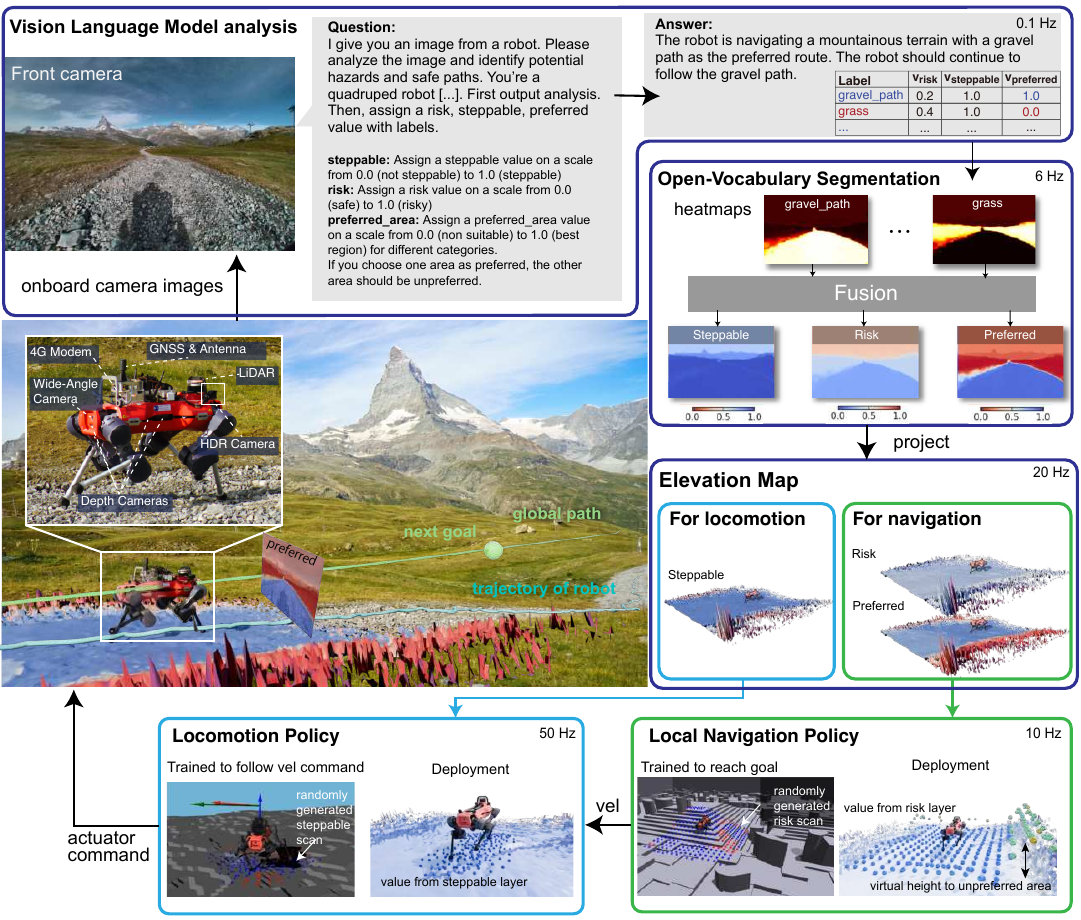} %
    \caption{\textbf{Overview of the ROAM pipeline.} The proposed system bridges high-level reasoning from vision-language models (VLMs) with low-level actuator commands for autonomous robot control. It processes RGB images and robot descriptions through a VLM to generate object labels enriched with risk, preferred area, and steppable values. These labels are converted into semantic images using open-vocabulary semantic segmentation, which operates at a higher frequency than the VLM analysis. The semantic images are then fused with elevation maps to create an environmental representation that reinforcement learning (RL) policies utilize for fast and robust robot control.} 
    \label{fig:result_overview} %
\end{figure*}

ROAM integrates a \ac{VLM} for high-level world knowledge with \ac{RL} policies for low-level control, bridging abstract scene understanding with concrete physical action. This allows the robot to interpret its environment, make informed decisions, and execute those decisions robustly on hardware. The overview of this system is shown in Fig.~\ref{fig:result_overview} and \href{https://youtu.be/xcqVpbgklD4}{Movie 1}.

Our perception system uses a \ac{VLM} to generate real-time, semantically-aware navigation costs. We provide the \ac{VLM} with RGB images, robot information, desired output value ranges (for example, a risk value from 0 to 1), and past prompts and outputs as context. The model then outputs semantic labels for objects in the scene, along with associated values, such as risk or preference. For example, it may assign a low risk to a gravel path, a moderate risk to the surrounding grass, and a high risk to water.
An open-vocabulary semantic segmentation module~\cite{xu2023side} then uses these VLM-generated labels to produce a dense, real-time risk image of the environment. This image is projected onto the robot's elevation map, providing the RL policies with high-frequency spatial information (from \SI{10}{\hertz} to \SI{50}{\hertz}) for responsive control. This abstraction enables RL policies to operate on a processed representation of the environment rather than raw sensor data, which simplifies the learning problem and sim-to-real transfer.

To convert this semantic information into motion, we train two distinct RL policies: one for low-level locomotion and one for local navigation.
The locomotion policy, an extension of prior work~\cite{miki2022wild,miki2024confined}, is responsible for stable movement and foothold selection, incorporating steppable area information into its observation to avoid placing feet on unsafe terrain.
The navigation policy, building on RL-based methods~\cite{lee2024learning}, handles local pathfinding, using the risk information from the perception system to avoid hazardous regions while navigating toward a goal.
Training these policies in the real world would be data-intensive and expose the robot to potentially damaging situations. Therefore, we train both policies extensively in simulation. Although simulation cannot perfectly capture real-world complexity, we sidestep this challenge by training the policies on abstract observations, such as height scans, which are easily randomized. This strategy allows the policies to learn robust, low-level behaviors without requiring a photorealistic simulation of all possible environments.

This hybrid architecture allows ROAM to address the limitations of previous approaches. By combining the common-sense reasoning of \acp{VLM} with the robustness of RL policies trained with domain randomization, our system eliminates the need for exhaustive real-world datasets or comprehensive simulations, enabling it to handle diverse and unstructured environments.

This semantic-spatial delegation, from semantic reasoning through low-level control, differs from recent VLM-based paradigms that require the foundation model to perform dense spatial reasoning by selecting from action candidates, outputting spatial waypoints, or generating low-level motor commands.

\section{Results}\label{sec:results}
\begin{figure*}[tbp] %
    \centering
    \includegraphics[width=\textwidth,height=0.78\textheight,keepaspectratio]{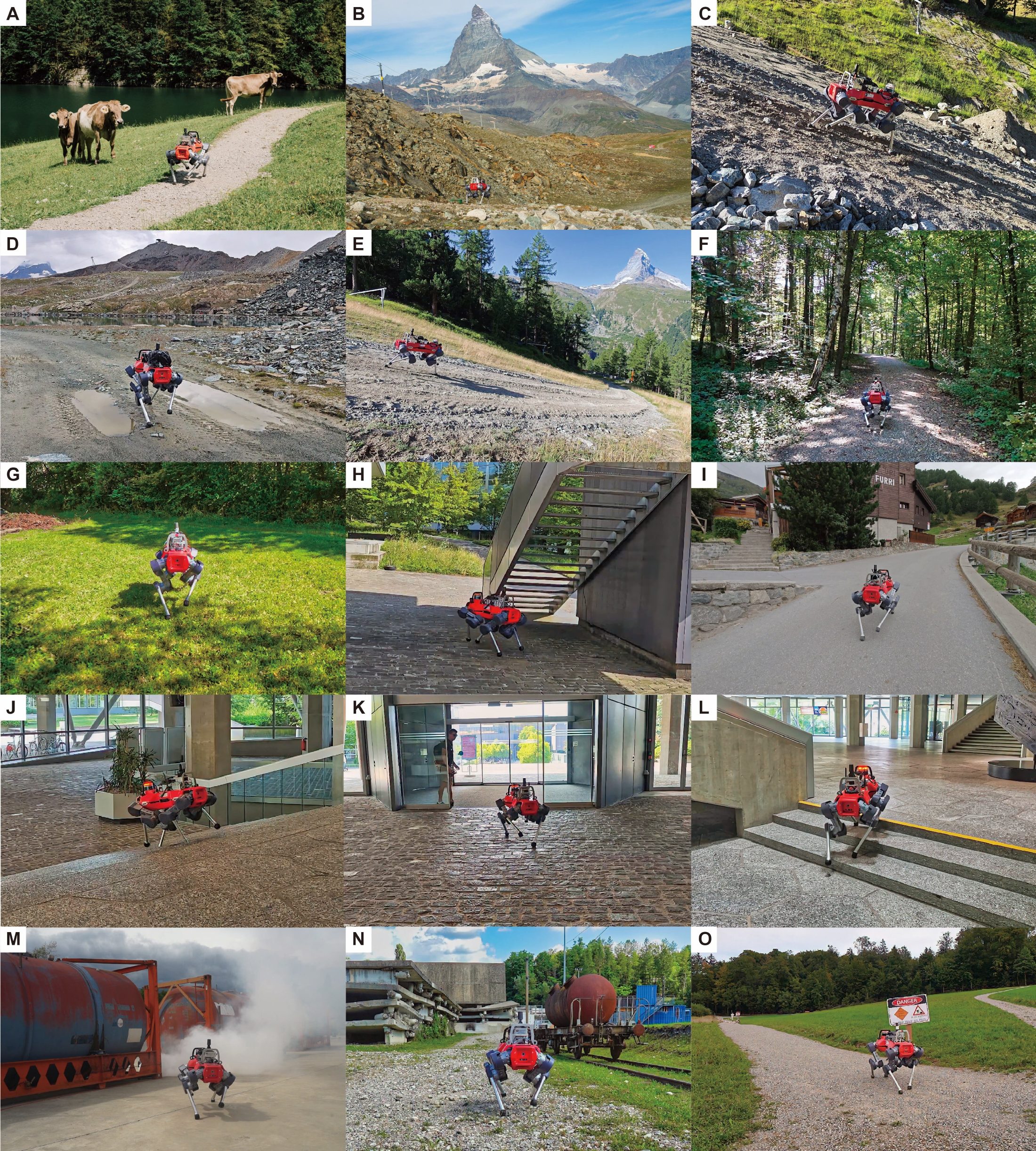} %
    \caption{\textbf{Deployment of ROAM across diverse environments.} These include natural outdoor settings such as mountain trails (A to E), forests (F), and grassy areas (G), as well as urban environments such as university campuses (H) and roads (I). The system was also tested in indoor settings (J to L). Additionally, we evaluated its performance in the presence of various challenges, such as smoke (M), train tracks (N), and signage (O).}
    \label{fig:result_various_envs} %
\end{figure*}

To evaluate the proposed ROAM system, we conducted experiments across diverse environments---including mountains, forests, urban areas, and indoor spaces---as shown in Fig.~\ref{fig:result_various_envs}. These experiments demonstrate that our method can generalize effectively to different settings without requiring data specific to those environments. The system assigns risks to different terrains, interprets signage, and adjusts its behavior accordingly, as illustrated in Fig.~\ref{fig:result_various_envs}O and \href{https://youtu.be/xcqVpbgklD4}{Movie 1}. 

\subsection{Hiking in the Swiss Alps}
\begin{figure*}[tbp]
    \vspace*{-1.8em}
    \centering
    \includegraphics[width=\textwidth,height=0.78\textheight,keepaspectratio]{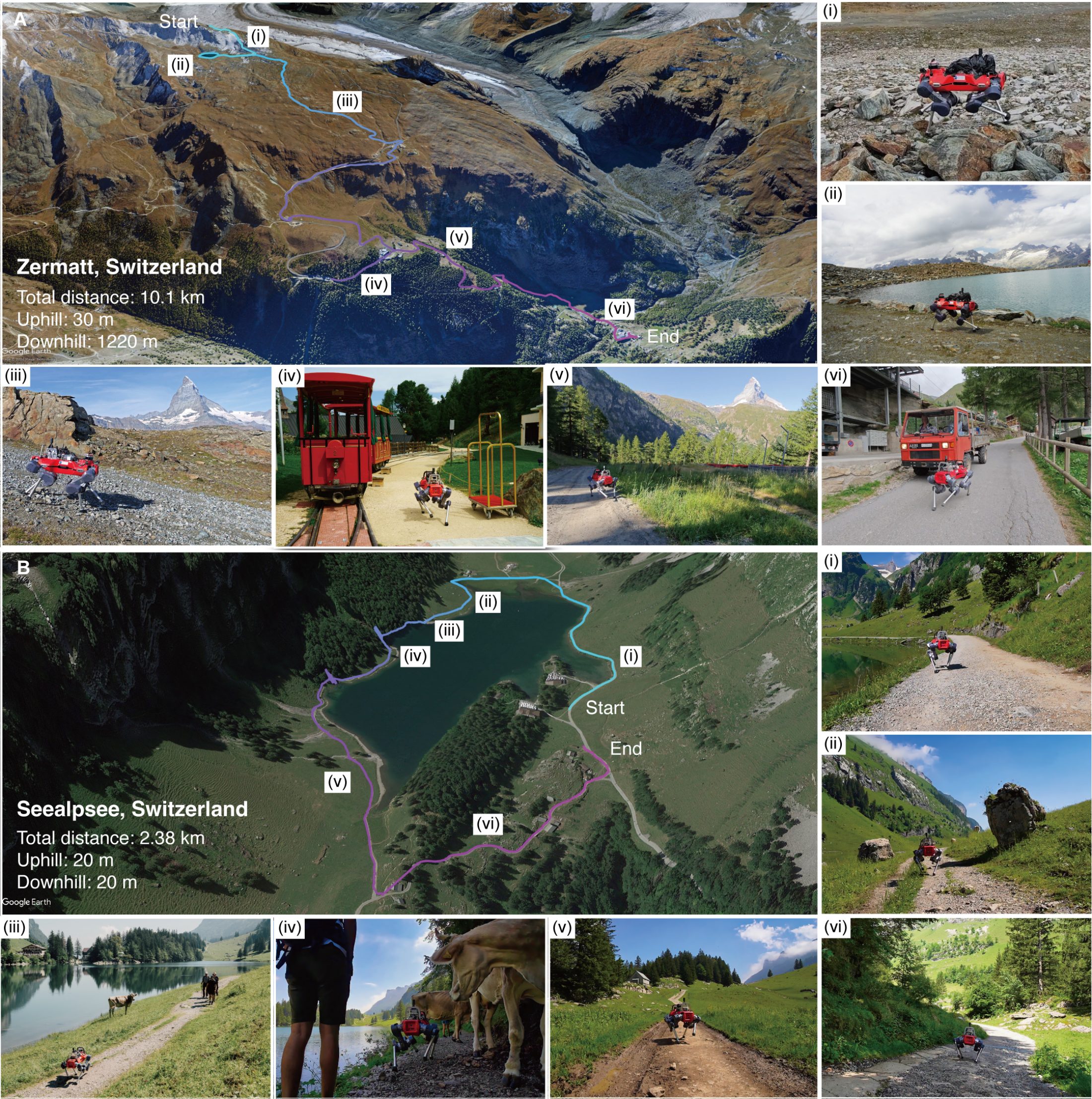} %
    \caption{\textbf{Long-distance hiking experiments in the Swiss Alps.} (A) Zermatt experiment. This experiment covered a total distance of \SI{10.1}{\kilo\meter} with an elevation change of \SI{1220}{\meter}. The route encompassed diverse terrains, including gravel paths, rocky sections, grassy areas, and segments near a lake and train tracks. The steepest segment of the hike had a gradient of approximately 30\%, posing a substantial challenge for locomotion.
(B) Seealpsee experiment. This experiment covered a total distance of \SI{2.38}{\kilo\meter} with an elevation change of \SI{20}{\meter}, following a path around the lake. The route included a narrow trail close to the water, encounters with cows, gravel sections, and a forest area near the goal. In (A) and (B), photographs (i) to (vi) show the robot at the correspondingly labeled locations on the map.}
    \label{fig:result_zermatt} %
\end{figure*}

To demonstrate the extent of autonomy we can achieve with our method, we conducted two long-distance hiking experiments in the Swiss Alps (Fig.~\ref{fig:result_zermatt} and \href{https://youtu.be/xcqVpbgklD4}{Movie 1}).

\subsubsection*{Zermatt}
This experiment demonstrated our robot's ability to navigate a varied and challenging hiking trail in the mountains. The route covered \SI{10.1}{\kilo\meter} with a \SI{1220}{\meter} elevation change, presenting different scenarios along the way.

The route started at the Gornergrat station (Fig.~\ref{fig:result_zermatt}A, Start), located over \SI{3000}{\meter} above sea level, and aimed to reach the Furi cable car station \SI{1220}{\meter} below (Fig.~\ref{fig:result_zermatt}A, End). This route presented several environmental challenges that required the robot to adapt its risk analysis.
The initial section consisted of gravel and small rocks, requiring robust locomotion over uneven terrain (Fig.~\ref{fig:result_zermatt}A(i)). As the robot progressed, it encountered a small lake, which required careful navigation along the water's edge without stepping into the lake (Fig.~\ref{fig:result_zermatt}A(ii)). This section highlighted the importance of semantic understanding from RGB images. After the lake, the robot followed a gravel path surrounded by rocks or partially high grass (Fig.~\ref{fig:result_zermatt}A(iii)). Here, semantic understanding helped to keep the robot on the safer gravel path, avoiding grass areas that can potentially hide steep slopes or rocks. Steep slopes with an approximately 30\% gradient also challenged the locomotion system (Fig.~\ref{fig:result_zermatt}A(iii)).
Near a station, the trail became more artificial, running along the railroad tracks (Fig.~\ref{fig:result_zermatt}A(iv)). After passing the mid-station, the path changed to a grassy trail in the forest (Fig.~\ref{fig:result_zermatt}A(v)). The last part of the route changed to an urban environment with asphalt ground near the cable car station (Fig.~\ref{fig:result_zermatt}A(vi)). Here, the system demonstrated its adaptability by adjusting labels, using terms such as ``road'' or ``paved\_road'' instead of ``gravel\_path,'' which helped segmentation.

Although the robot walked the majority of this route fully autonomously, two major human interventions were necessary to ensure safety due to the presence of construction zones on the hiking trail. In the first case, the fence in front of the path leading to the construction zone was not correctly segmented, even though the \ac{VLM} identified a fence. As a result, the robot walked into it. The second intervention became necessary when a deviation in the global navigation path would have led the robot into hazardous construction terrain. In this case, the segmentation model was unable to adequately differentiate between the designated path and the rough construction area, necessitating human correction to prevent the robot from entering the dangerous zone. Both interventions were necessary to maintain the robot's safety (fig.~\ref{fig:internvention}).

Despite these challenges, the robot completed most of the hike autonomously without human assistance. The autonomy rate in time was 96.7\% throughout the entire mission. This experiment demonstrated the effectiveness of our integrated approach.

\subsubsection*{Seealpsee}
In our second experiment, we tested our system on a trail around the lake Seealpsee in Switzerland. This \SI{2.38}{\kilo\meter} course presented a variety of challenges, including narrow paths adjacent to the lake (Fig.~\ref{fig:result_zermatt}B(i, iii, and iv)), pastures (Fig.~\ref{fig:result_zermatt}B(i to iii and v)), trees impairing \ac{GNSS} signal (Fig.~\ref{fig:result_zermatt}B(vi)) and the presence of animals such as cows (Fig.~\ref{fig:result_zermatt}B(iii and iv)). The complexity of the environment highlighted the limitations of purely geometry-based navigation and showed the need for semantic understanding. There were several sections near the lake where the global path deviated from the actual local path and misdirected the robot into the water (Fig.~\ref{fig:result_zermatt}B(i)). In addition, the presence of pastures along the path  required adherence to semantic norms. Although it is safe to cross them, it is undesirable to do so since they are private property owned by farmers.

Our system successfully prevented the robot from entering the lake or restricted pastures by using semantic information from the environment, recognizing the boundaries of the lake and the pastures, and adjusting risk and preference values accordingly to stay on the designated path. We also observed that the system's output reflects the context: it only identified ``water'' or ``water\_body'' when the lake was visible, detected increased risk from trees when entering wooded areas, and adjusted risk values for areas with dense vegetation (fig.~\ref{fig:seealpsee-labels}).
Although the robot's navigation was largely autonomous, several operator interventions were necessary to guarantee safety. One such instance involved halting the robot as it approached the edge of a lake. Another required intervention was at a turnstile, which presented an impassable physical barrier for the robot. Despite these exceptions, the system achieved a 97.5\% autonomy rate for the mission.

\subsection{Contextual awareness and sign comprehension}

\begin{figure*}[tbp]
    \vspace*{-1.8em}
    \centering
    \includegraphics[width=\textwidth,height=0.78\textheight,keepaspectratio]{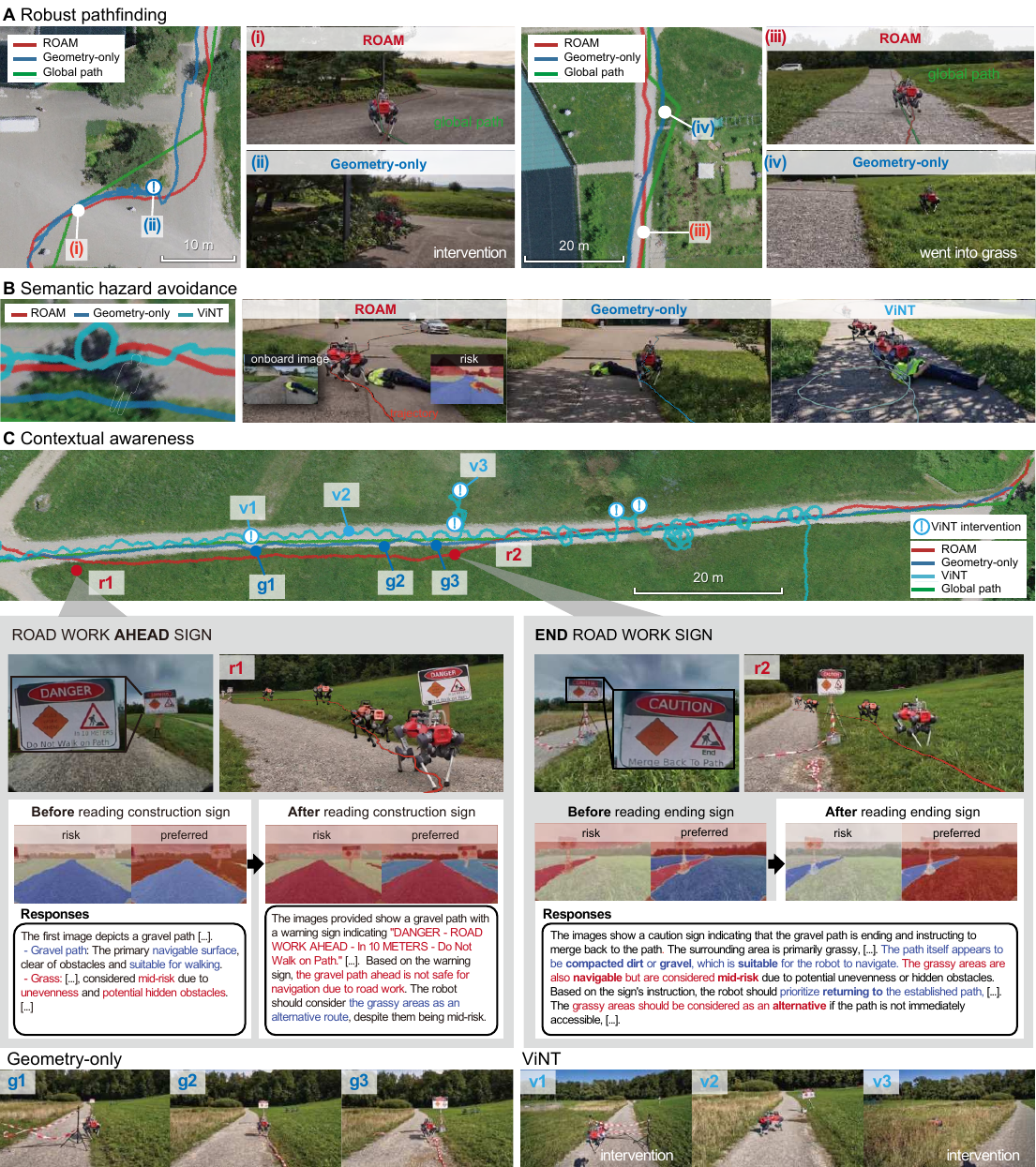} %
    \caption{\textbf{Comparison of ROAM with baselines on a course with semantic and contextual challenges.} The experiment covered a \SI{1.2}{\kilo\meter} path with challenges such as an inaccurate global path deviating into bushes or grass, a mannequin lying on the path as an obstacle, and a construction zone with warning signs. (A) Robust pathfinding. The global path was offset into bushes [(i) and (ii)] and into grass [(iii) and (iv)]; (i) and (iii) show ROAM, and (ii) and (iv) show the geometry-only baseline. (B) Semantic hazard avoidance with a mannequin lying on the path for ROAM, the geometry-only baseline, and ViNT. (C) Contextual awareness with construction signs. The map shows the paths of the three methods and the global path. The remaining panels show the robot views, the risk and preferred-area images, and the VLM responses before and after ROAM read the ``ROAD WORK AHEAD'' sign (r1) and the ``END ROAD WORK'' sign (r2), followed by snapshots of the geometry-only baseline (g1 to g3) and ViNT (v1 to v3). Information icons mark the ViNT interventions.} 
    \label{fig:result_construction} %
\end{figure*}

We designed a controlled experiment consisting of several challenges along a \SI{1.2}{\kilo\meter} course at the H\"onggerberg campus in Zurich, Switzerland, to evaluate the behavior of our system in the real world, on the real robot (Fig.~\ref{fig:result_construction} and \href{https://youtu.be/xcqVpbgklD4}{Movie 1}). 

We compared the ROAM system with two baseline methods: the Field Local Planner~\cite{fieldlocalplanner2022}, which relies solely on geometric information without explorative behavior, and Visual Navigation Transformer (ViNT)~\cite{shah2023vint}, which uses goal images as navigation commands. To obtain these goal images for ViNT, we first teleoperate the robot through the course and then use images sampled every \SI{1.5}{\second} as navigation goals.
We describe the challenges and how our method performs in comparison to the baselines in the following sections.

\subsubsection*{Robust pathfinding}
This section evaluates the system's robustness to deviations between the planned path and the environment, introduced via artificial offsets that mimic naturally occurring discrepancies arising from inconsistent global path data and global localization drift. The results, visualized in Fig.~\ref{fig:result_construction}A, demonstrate that ROAM seamlessly handles such discrepancies. For instance, when the global path passed through a patch of bushes (Fig.~\ref{fig:result_construction}A(i)), ROAM identified an alternative route. This was achieved through its local navigation policy and the semantic guidance derived from the VLM's segmentation. In contrast, the geometry-only method (Fig.~\ref{fig:result_construction}A(ii)) proceeded along the commanded path into the bushes. The robot, unable to reconcile the global waypoint with the high geometric cost of the terrain, became stuck and required human intervention to continue the experiment.
We repeated the experiment in a grassy area, this time without any geometric obstacles. Our method stayed on the gravel path (Fig.~\ref{fig:result_construction}A(iii)) thanks to the ``preferred\_area'' estimation, which favored the designated path and avoided the grass. The geometry-only baseline blindly followed the global path and traversed the grass due to its lack of semantic understanding (Fig.~\ref{fig:result_construction}A(iv)). 

\subsubsection*{Semantic hazard avoidance}
We demonstrate the common-sense reasoning capability of our system by placing a mannequin on the path and reporting the resulting behavior.  Rather than representing an immediate physical or geometric hazard, the mannequin serves to test the VLM's ability to contextually identify a human figure as a high-risk obstacle. As shown in Fig.~\ref{fig:result_construction}B, our method identified the mannequin as an obstacle with the label ``human.'' The local navigation policy used this semantic information and avoided it. In contrast, the geometry-only method, lacking semantic scene understanding, failed to recognize the mannequin as an obstacle and stepped on its leg. The ViNT method struggled to stay on the path, requiring multiple human interventions, and did not change its behavior in the presence of the mannequin, eventually stepping on it as well.

\subsubsection*{Contextual awareness}
To assess the robot's ability to interpret and respond to contextual information, we placed construction signs along the path. Upon encountering the first sign, which read, ``DANGER - ROAD WORK AHEAD IN 10 METERS - Do not walk on path,'' ROAM correctly interpreted the sign and adjusted its risk assessments, marking the gravel path as high risk and the grass as a preferred route. As a result, the robot correctly took a detour through the grass.
After navigating through the grass, the robot encountered a second sign that read, ``CAUTION - END ROAD WORK - Merge Back to Path.'' ROAM correctly interpreted this sign, updating its evaluations and marking the gravel path as preferred and low risk. The robot then returned to the path and continued its journey (Fig.~\ref{fig:result_construction}C).

The geometry-only baseline failed to interpret the signs and continued to follow the path based on the global waypoints, resulting in the robot entering the construction site. The ViNT method also failed to read the signs and instead followed the path based on the pre-captured target images. However, these images did not include the construction site, and ViNT was unable to adapt to this change in the environment. As a result, it failed to avoid the barrier, collided with it, and deviated into tall grass several times, requiring intervention. Ultimately, ViNT was unable to recover and continued through the grass in the wrong direction. 

\subsection{Validation of system components}
To further validate our approach, we conducted several experiments showcasing how ROAM applies common-sense reasoning to locomotion and local navigation. We specifically demonstrate how semantic information is used to step only in safe areas, adjust path risk based on semantic understanding, and further illustrate the exploration behavior.

\subsubsection*{Locomotion}
\begin{figure*}[tbp]
    \vspace*{-1.8em}
    \centering
    \includegraphics[width=\textwidth,height=0.78\textheight,keepaspectratio]{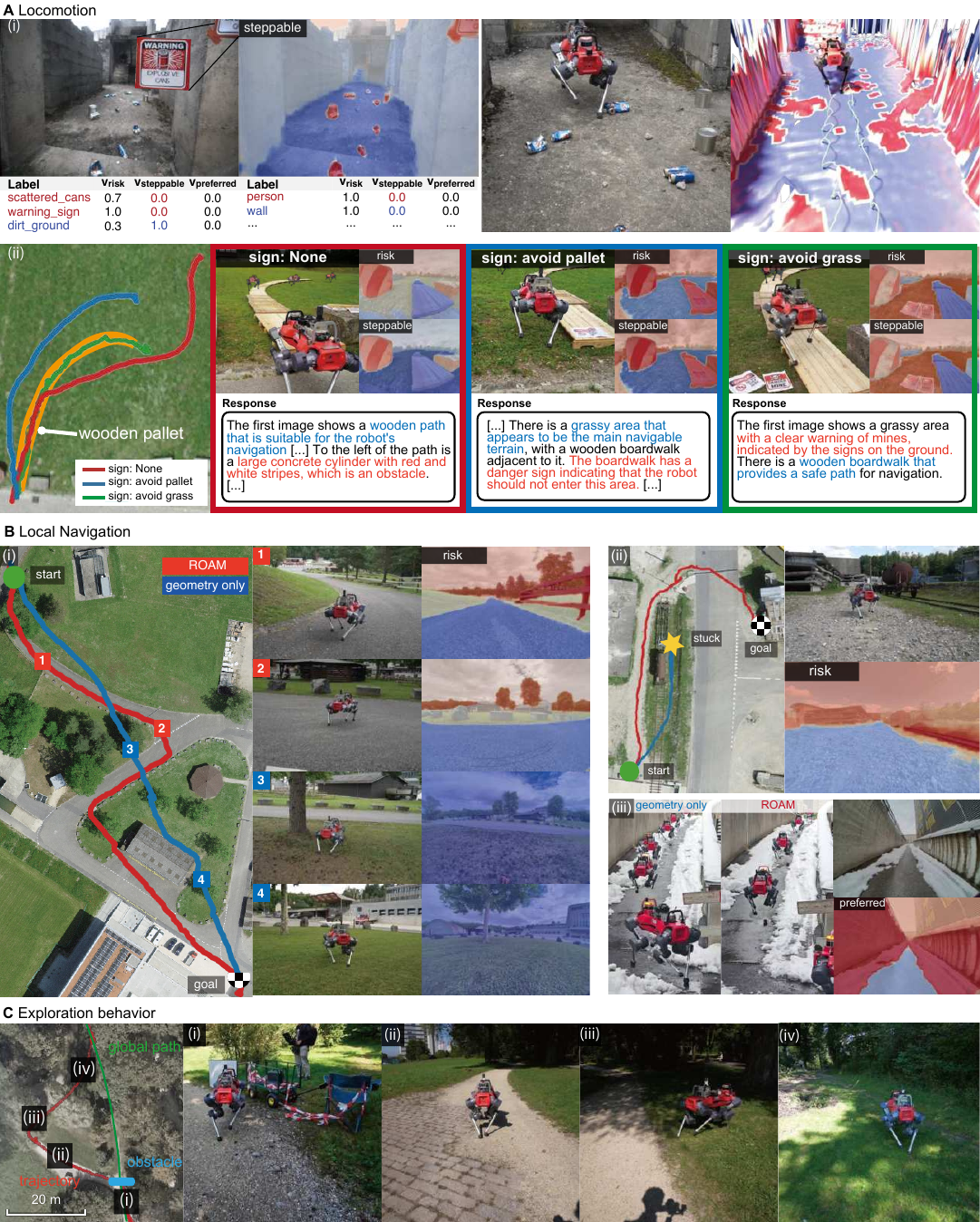} %
    \caption{\textbf{ROAM brings world knowledge to locomotion, local navigation, and exploration.} (A) ROAM reads signs and uses this information in locomotion: (i) metal cans with a ``WARNING: EXPLOSIVE CANS'' sign and (ii) wooden pallets on grass without a sign, with a sign on the pallet, and with signs on the grass. The table in (i) lists the VLM labels and their values, the boxes in (ii) show the VLM responses, and the map in (ii) shows the wooden pallet (orange outline) and the foot trajectories for the three sign conditions. (B) Comparison of ROAM with a geometry-only baseline on (i) a campus route, (ii) a route next to a railroad, and (iii) a snow-covered path. In (i), images 1 to 4 correspond to the numbered points along the paths taken by the two methods, highlighting the different terrains encountered, such as paved areas, gravel, and grass. (C) Exploration behavior in the presence of an obstacle blocking the path: (i) the blocked path, (ii) the alternative route along the path, (iii) the shortcut through the grass, and (iv) the return to the global path. The robot explored along the path after encountering the obstacle and took a shortcut through the grass to merge back to the global path.}
    \label{fig:arche} %
\end{figure*}

To test ROAM's ability to alter foot placement based on context, we placed metal cans on the ground and added a warning sign reading ``WARNING: EXPLOSIVE CANS'' (Fig.~\ref{fig:arche}A(i)). The \ac{VLM} successfully interpreted the sign, assigning a high-risk value to the cans and marking them as non-steppable. The locomotion policy used the contextual information to avoid stepping on the cans, selecting safe footholds based on the steppable layer of the elevation map.

Next, we placed wooden pallets to create a path across the grass (Fig.~\ref{fig:arche}A(ii)). Without any signs, the VLM evaluated both the pallet and the grass as low-risk areas, with the pallet being slightly less risky. As a result, the robot traversed both the pallet and the grass to reach the goal. When we placed a ``DO NOT STEP HERE'' sign on the wooden pallet, the VLM updated its evaluation, marking the pallet as high-risk and non-steppable. The robot then adapted its behavior, choosing to walk only on the grass to reach the goal. Conversely, when we placed both ``DO NOT STEP HERE'' and ``DANGER: MINE'' signs on the grass, the VLM marked the grass as high-risk and non-steppable. In response, the robot stepped only on the wooden pallet. This demonstrates how our system can adapt to different contextual cues, modifying behavior in response to changes in the environment, even within the same physical space.

\subsubsection*{Local navigation}
We conducted a comparison of the paths taken by the robot with and without semantic information. In the setting without semantic information, all risk values were set to 0.0.
Our local navigation policy combines geometric data with a semantic risk map to determine a path. This adjustment effectively disabled the semantic input, ensuring the policy relied solely on geometric information for local navigation.
We set a local navigation goal about \SI{200}{\meter} away and allowed the local navigation policy to reach this goal. With semantic information, the robot preferred to walk on the road. In contrast, when no semantic information was used, and only geometric information was provided, the robot chose to move directly towards the goal by walking on the grass (Fig.~\ref{fig:arche}B(i)). In addition, when the robot was near a railroad, our method prevented it from entering the rail area. Without semantic information, the robot crossed the rail and ended up in a dead-end area. These results indicate that by using the VLM and semantic segmentation, our method can produce more socially preferred paths compared to cutting across grass and rails (Fig.~\ref{fig:arche}B(ii)).
Furthermore, we conducted an experiment on a snow-covered path where the route was surrounded by snow. Our method successfully identified the snow-covered areas as less preferable and remained on the asphalt path. In contrast, the geometric baseline did not adjust its behavior in response to the presence of snow and traversed the snow-covered area instead (Fig.~\ref{fig:arche}B(iii)).

\subsubsection*{Exploration}
In this experiment, we place barriers that block the intended path, forcing the robot to explore alternative routes (Fig.~\ref{fig:arche}C(i)). ROAM recognizes the blockade and chooses an alternative path (Fig.~\ref{fig:arche}C(ii)). When the path deviates from the goal, the robot cuts through the grass (Fig.~\ref{fig:arche}C(iii)) and rejoins the main path (Fig.~\ref{fig:arche}C(iv)). This was possible because the grass area was classified as not preferred, but not high risk.

\subsection{Evaluating different levels of action abstraction}
\begin{figure*}[tbp]
    \vspace*{-1.8em}
    \centering
    \includegraphics[width=\textwidth,height=0.78\textheight,keepaspectratio]{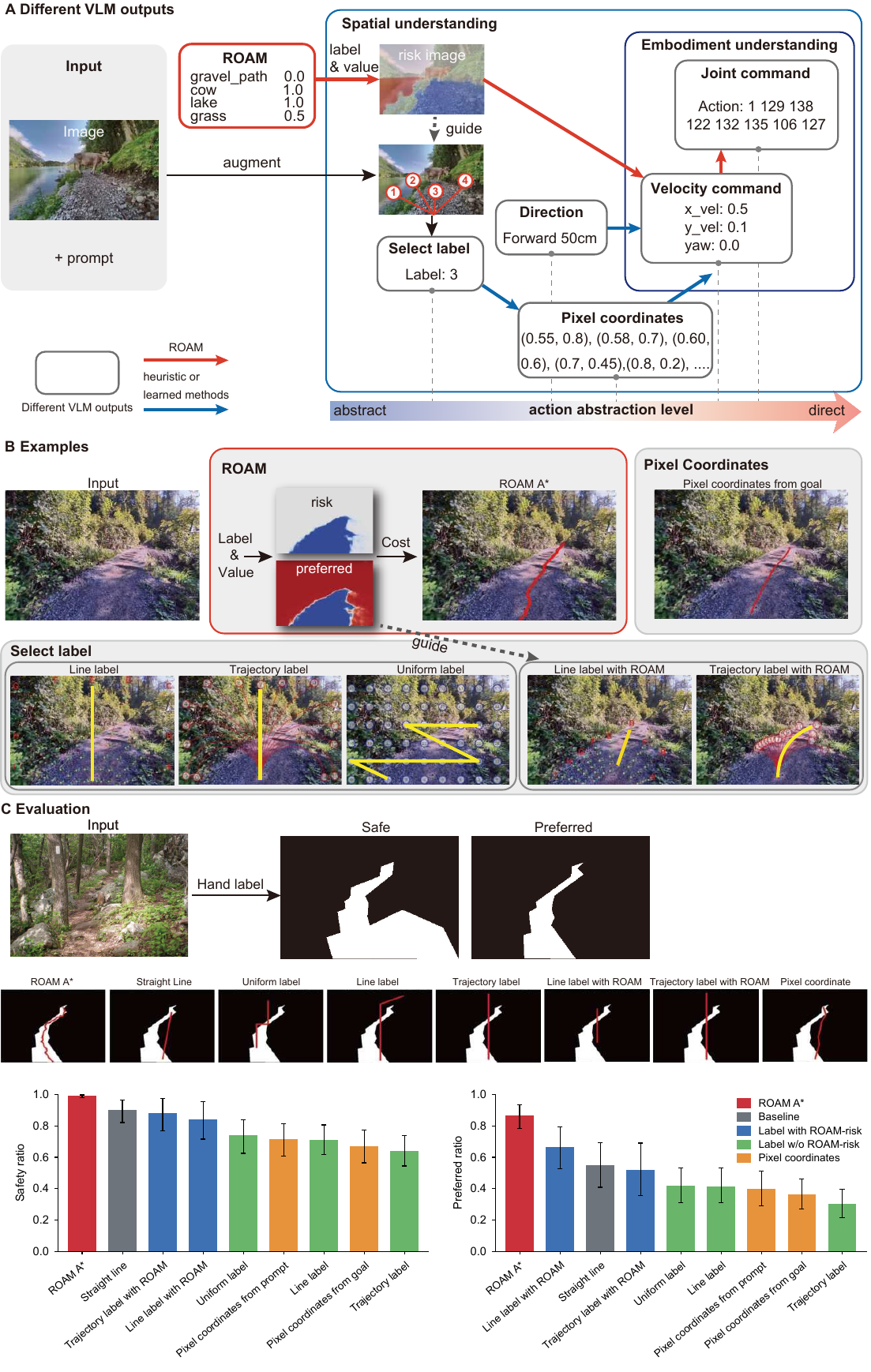} %
    \caption{
    \textbf{Comparison of VLM output abstractions for navigation.} (A) Comparison of VLM-based navigation methods by output type. ROAM outputs semantic labels and values, avoiding the need for spatial understanding. Methods outputting directions or coordinates require spatial reasoning; those generating direct commands (such as velocity) additionally require embodiment understanding. (B) Example trajectories from different methods, including three ``select label'' variants. ROAM information can augment label placement on ground regions. (C) Quantitative evaluation of the safety and preferred ratios, defined as the fraction of all trajectory pixels, pooled over the scenes (N = 27 scenes; see Statistical analysis), that lie within the hand-labeled ``safe'' and ``preferred'' regions. Error bars, 95\% bootstrap confidence intervals.
    }
    \label{fig:vlm} %
\end{figure*}

We evaluate our method by comparing different action spaces for \ac{VLM}-based navigation. The choice of action space presents different levels of abstraction and different requirements for spatial and embodiment understanding. We illustrate this in Fig.~\ref{fig:vlm}A. Although any high-level command must ultimately be translated into high-frequency joint-level actions (typically \SIrange{50}{100}{\hertz}), a key differentiator is where spatial reasoning occurs.

Our proposed method, which outputs semantic labels and values, delegates spatial processing to downstream modules. This design frees the VLM from needing intrinsic spatial understanding. Conversely, alternative approaches, such as direct trajectory pixel coordinate generation, necessitate that the VLM itself performs spatial reasoning to produce geometrically correct outputs.
More extreme methods, such as \ac{VLA} models, attempt to directly generate low-level motor commands. They place a greater burden on the model by requiring not only spatial reasoning but also a deep embodied understanding of the robot's kinematics and dynamics. Furthermore, these models often require extensive fine-tuning and must meet the high-frequency demands of dynamic locomotion.

In this initial investigation, we focus on methods that leverage pre-trained VLMs in a zero-shot capacity, using prompting to elicit navigational intelligence. We evaluate several distinct action spaces, ranging from direct trajectory pixel coordinate generation to the semantic risk-based planning used in ROAM.

\subsubsection*{Experimental setup for path generation}
To provide a controlled comparison, we evaluate the ability of a \ac{VLM} to generate a safe and viable path to a designated goal within a static image. For our method, ROAM, we replace the full navigation policy with a standard A* path planner to ensure the comparison focuses purely on the path generation in the image space. 

We make this methodological choice to decouple the semantic-spatial perception quality from the robot's physical control dynamics. Deploying the trained RL navigation policy is incompatible with a static image dataset, since it requires a continuous stream of real-time proprioceptive states and a dynamic 2.5D elevation map. Furthermore, because the baseline methods we evaluate against predict paths as pixel coordinates within the 2D image space, an A* planner operating in the same 2D image space provides a direct proxy for comparison.

In this setup, the planner uses the semantic risk map as its cost function. We benchmark this approach against several baseline methods that use the VLM to generate a path more directly.

We compared three types of methods. In our proposed method, ROAM, the VLM is prompted to output a list of semantic labels present in the scene (such as ``pavement,'' ``tall grass,'' and ``obstacle'') along with their associated navigation risk (a value from 0 to 1). These text-based outputs then guide an open-vocabulary segmentation model to generate a dense risk map. The A* algorithm planned a least-cost path through this map to the same goal coordinate used in the baseline methods.

In the direct trajectory pixel coordinate baseline, the VLM is prompted with a natural language command (such as ``Generate a trajectory that follows the path'') or a goal coordinate and is asked to output a sequence of pixel coordinates forming a path (similar to that of Windecker et al.~\cite{windecker2025navitrace}).

In the label-based path selection baseline, we overlay discrete labels onto the image and prompt the VLM to output a sequence of these labels to form a path toward the goal. We explored several labeling strategies, placing labels along straight lines with ticks, simple trajectories, or distributing them in a grid. To assess the impact of prior terrain knowledge, we tested two variants for label placement: distributing them over the entire image versus placing them only on navigable ground, which was identified using the steppable map from our own ROAM method.

\subsubsection*{Qualitative analysis}
Fig.~\ref{fig:vlm}B presents a qualitative comparison of the paths generated by each method on a representative scene. The results highlight a notable challenge for current VLMs: a difficulty with precise spatial reasoning. Methods relying on the direct output of coordinates or selecting labels produced erratic or unfeasible paths that fail to respect clear environmental boundaries. For instance, the direct trajectory prompts often result in paths that cut across unsafe terrain.

In contrast, the path generated via the ROAM pipeline is visibly more reliable and coherent. By tasking the VLM with semantic understanding, such as identifying path, grass, water, and their risks, and offloading the spatial task to semantic segmentation and geometric path planning by A*, our method successfully leverages the VLM's strengths in common-sense reasoning. The VLM does not need to be spatially precise; it only needs to correctly identify the concepts in the scene. This allows the subsequent, spatially-aware planning module to generate a safe and logical trajectory. 

\subsubsection*{Quantitative analysis}
To quantify these observations, we conducted an analysis across 27 outdoor and indoor scenes (see Statistical analysis). For each image, we manually segmented ``safe'' and ``preferred'' regions for traversal. We then measured the fraction of all trajectory pixels, pooled over the scenes, that lay within the ``safe'' regions and ``preferred'' regions. A trajectory was considered a ``total failure'' if it entered an unsafe region, as this could correspond to a mission-ending event for a real robot, whereas the ``preferred'' percentage shows how much the trajectory follows the human preference.

The results, summarized in Fig.~\ref{fig:vlm}C and Table~\ref{tab:vlm-comparison}, demonstrate that the trajectories generated using the ROAM pipeline align better with the human labeler's intention, outperforming direct action generation baselines. Our method consistently produces paths that remain within safe regions, reflecting its robustness. This suggests that for the current generation of foundation models, abstracting the VLM's role to a high-level semantic advisor is a more effective strategy for navigation than tasking it with generating concrete, low-level robot actions, which requires spatial reasoning.

\begin{table*}[tbp]
\centering
\small
\setlength{\tabcolsep}{4pt}
\begin{tabular}{l l S[table-format=1.3] c S[table-format=1.3] c c}
\toprule
Method & Category & {Safety} & \textit{P} & {Preferred} & \textit{P} & \shortstack{Goal\\reached} \\
\midrule
ROAM A* & ROAM A* & {\textbf{0.991}} & -- & {\textbf{0.868}} & -- & 27 \\
Straight line & Baseline & {0.900} & 0.037 & {0.550} & 0.001 & 27 \\
Trajectory label with ROAM & Label with ROAM-risk & {0.880} & 0.037 & {0.519} & 0.006 & 3 \\
Line label with ROAM & Label with ROAM-risk & {0.840} & 0.055 & {0.668} & 0.037 & 5 \\
Uniform label & Label w/o ROAM-risk & 0.740 & $<$0.001 & 0.418 & $<$0.001 & 4 \\
Pixel coordinates from prompt & Pixel coordinates & {0.714} & 0.001 & {0.398} & $<$0.001 & 5 \\
Line label & Label w/o ROAM-risk & {0.711} & 0.002 & {0.414} & $<$0.001 & 1 \\
Pixel coordinates from goal & Pixel coordinates & {0.670} & $<$0.001 & {0.363} & $<$0.001 & 3 \\
Trajectory label & Label w/o ROAM-risk & {0.639} & $<$0.001 & {0.302} & $<$0.001 & 3 \\
\bottomrule
\end{tabular}
\caption{\textbf{Safety and preferred ratios of VLM-based path generation methods.} Values are fractions of trajectory pixels pooled across N = 27 hand-labeled scenes (see Statistical analysis). Goal reached denotes scenes ending within 20\% of the start-to-goal distance from the goal; \textit{P}, Holm-corrected two-sided Wilcoxon signed-rank test against ROAM A*.}
\label{tab:vlm-comparison}
\end{table*}

\section{Discussion}
In this work, we have presented ROAM, an autonomous navigation system that bridges high-level common-sense reasoning from \acp{VLM} with learned control policies to guide legged robots across complex real-world environments. The core of our approach is a framework that translates the abstract, common-sense reasoning of \ac{VLM}s into actionable, low-level control. Our approach aims to address the poor spatial awareness in VLM-based robotics. Our method circumvents the inaccuracies of directly querying \ac{VLM}s for coordinates or commands by instead using them for generating labels for open-vocabulary semantic segmentation. This semantic information is then projected onto a spatial map, which acts as a bridge to our \ac{RL} navigation and locomotion policies. This strategy distinguishes our work from methods that rely solely on geometric data or require costly, environment-specific retraining. As a result, our system exhibits strong zero-shot generalization, navigating effectively across various domains, including forest trails, rocky inclines, and indoor spaces, without requiring fine-tuning.

The first advantage of our approach is the ability to apply common sense, specifically interpreting semantic information and environmental context without the need for hard-coded labels or rules for every possible scenario. For example, the robot recognized and avoided risky areas such as a lake, respected private pastures, and used textual information on signage to dynamically adjust traversal strategies. 
This common understanding is critical in situations where geometric information is insufficient or misleading. The hiking experiments in the Swiss Alps, which covered kilometer-scale distances, demonstrate how our system handles different terrain variations.

The second advantage of our method is to fuse the high-level, common-sense reasoning of VLMs, which can operate at slower frame rates and tolerate larger processing delays, with the reactive, low-latency control outputs of RL-based locomotion and navigation policies. Unlike methods that struggle with latency-sensitive tasks, our hierarchical integration ensures that high-level semantic reasoning does not compromise rapid foot placement and local navigation.

However, several limitations and areas for improvement remain. The quality of semantic interpretation depends on both the accuracy of the large vision-language models and the reliability of the open-vocabulary segmentation. In practice, these models may occasionally misinterpret complex scenes, and the segmentation may not match the semantic regions identified by the VLM. For example, although the VLM might label an area as ``left\_grass'' or ``nearby\_rocks,'' the segmentation process does not accurately segment these corresponding regions. 

Also, engineering solutions like the sign detector were added to address the current VLM inference latency. As VLM speeds improve, these temporary solutions should become obsolete.

A fundamental characteristic of ROAM is that its common-sense reasoning is inherited from a pretrained VLM rather than learned onboard. Consequently, the system's ability to interpret signs, recognize hazards, and apply contextual norms is bounded by the capabilities of the underlying foundation model. Although our field experiments relied on a commercial cloud-hosted VLM, which introduces dependencies on network connectivity, external service availability, and closed-source model weights, ROAM is deliberately designed as a model-agnostic framework.

The gap between commercial and open-weight VLMs is closing and the performance of smaller open-weight VLMs is improving rapidly. As detailed in our supplementary VLM benchmark, current open-weight models already achieve highly competitive performance on zero-shot terrain classification. Although they currently trail leading commercial models in tasks requiring complex contextual grounding, such as precisely relating textual sign instructions to specific physical objects, we expect this gap to diminish quickly. Once onboard open-weight VLMs become capable enough to serve as the primary reasoning component, they will remove the network dependency and make the system fully self-contained. The architectural pattern introduced in ROAM, where a VLM produces semantic labels with continuous risk and preference values that are subsequently projected into spatial maps and consumed by RL policies, is designed to accommodate any sufficiently capable VLM, positioning the system to benefit directly from continuous progress in open foundation models.

Furthermore, our primary field experiments were conducted under daylight conditions with generally favorable visibility. We acknowledge that the system's performance under extreme visual degradation, such as heavy fog, heavy rain, night, or strong backlight, was not tested extensively. However, a key advantage of ROAM's hybrid architecture is its graceful degradation. Since the low-level locomotion and local navigation policies also rely on active sensors (such as LiDAR and stereo depth cameras) to build a geometric terrain map, the system can safely fall back to purely geometric navigation if the VLM's semantic visual inputs are completely compromised.

Additionally, our current system depends on a global reference path for long-range missions, as the VLM lacked the capability to interpret global maps sufficiently to provide reliable route planning based solely on onboard imagery.

Another limitation is that our current semantic-spatial abstraction primarily captures visual hazards and lacks the granularity to convey fine-grained material properties. Distinguishing the subtle risk differences between thin ice, deep mud, or loose gravel relies heavily on physical interaction rather than purely visual semantics. Although ROAM currently mitigates this at the control level---our RL locomotion policy is trained with extensive domain randomization and is robust to such physical disturbances---the system's high-level ``common sense'' remains disconnected from actual physical contact. Current typical VLMs operate purely on visual and textual inputs and do not accept proprioceptive or tactile force information. Enabling the next generation of foundational models to natively process and reason over this type of rich, embodied physical interaction data represents a crucial next step for the field.

In summary, this work takes a step toward unifying high-level semantic understanding with agile, low-level control for legged robots. However, substantial challenges remain. Although \ac{VLM}s demonstrate considerable potential, our findings suggest that further advances are needed before they can enable robots to navigate ubiquitously and autonomously in the complexities of the diverse real world.

\section{Materials and Methods}

\subsection{Study design}
The objective of this study was to evaluate whether the common-sense reasoning of a pretrained \ac{VLM}, combined with locomotion and local navigation policies trained with \ac{RL}, enables a legged robot to navigate autonomously in unstructured real-world environments. We conducted two long-distance hiking deployments in the Swiss Alps (Zermatt, \SI{10.1}{\kilo\meter}; Seealpsee, \SI{2.38}{\kilo\meter}), in which we recorded the operator interventions, their causes, and the autonomy rate. On a controlled \SI{1.2}{\kilo\meter} course, we compared ROAM with a geometry-only baseline~\cite{fieldlocalplanner2022} and ViNT~\cite{shah2023vint} under artificial path offsets, a mannequin on the path, and construction signs. Further real-world experiments isolated the effect of semantic information on locomotion, local navigation, and exploration. In an image-based evaluation on 30 hand-labeled scenes, we compared VLM output abstractions in terms of the fraction of the trajectory pixels that stayed within safe and preferred regions. Supplementary experiments benchmarked several VLMs, ablated the prompt, and tested the reasoning of VLMs about stair height. Each hiking deployment and each controlled-course run was performed once per method (N = 1), and the mannequin avoidance experiment was repeated 10 times per condition (N = 10). Sample sizes were determined by the available field time and the size of the labeled datasets rather than by a power analysis. In the image-based evaluation, the label-selection pipeline produced no output for 3 of the 30 scenes; these scenes were therefore excluded for all methods, and all methods were compared on the remaining 27 scenes. No field runs or mannequin trials were excluded.

\subsection{Hardware configuration}
We used the ANYmal D quadruped robot as our platform and customized it with sensors to meet the specific requirements of our mission. The robot is equipped with six RealSense stereo cameras for close-range depth perception and a Velodyne LiDAR sensor mounted on top for long-range depth measurements. In addition, we integrated a front-mounted C1 camera with a 198-degree field of view and high dynamic range for improved visibility in varying lighting conditions. A default wide-angle rear camera provides visual coverage of the rear.
For global localization, we equipped the robot with a Swift Navigation Piksi Multi \ac{GNSS} module using Real-Time Kinematic (RTK) corrections from the Swiss Federal Office of Topography (swisstopo).
To maintain internet connectivity during field operations, we integrated a Teltonika RUTX11 router with a SIM modem to enable real-time data transfer and processing. This connectivity supports communication with external systems, including OpenAI's API for the GPT-4 vision model. To increase the robot's computing capacity, especially for processing camera data and generating elevation maps, we added an NVIDIA Jetson AGX Orin module.

\subsection{Software overview}
To enable autonomous navigation, our system is composed of multiple modules, ranging from low-level control to high-level navigation. These components operate as independent nodes within the Robot Operating System (ROS) and communicate through various topics. 

Each autonomous mission starts with the global graphical user interface (GUI) module, where the user initiates the mission by selecting coarse waypoints on a loaded satellite image. The module then generates an interpolated path between these rough waypoints using a server based on OpenRouteService. This global path is represented in longitude-latitude-altitude (LLA) coordinates and requires a transformation to the local robot frame for further navigation.
To achieve this transformation, we estimate the robot's pose by fusing data from the LiDAR point cloud with \ac{GNSS} measurements. 

We sample local lookahead goals from the converted path, which are fed into the local navigation policy. The local navigation module outputs the velocity command to reach the given goal and sends it to the locomotion module, which controls each actuator in the robot's legs to perform the movement.
In parallel, a local elevation mapping module runs to continuously fuse LiDAR and depth-camera point cloud data with semantic information such as risk or steppable areas. These semantic labels are generated by a separate semantic segmentation module that queries the \ac{VLM} for new labels to improve environmental understanding.
Details of each method are explained in the following sections.

\subsection{Environmental perception}
Environmental perception is achieved through the fusion of data from RGB cameras, depth cameras, and LiDAR sensors. 

\subsubsection*{Vision-language model--based scene evaluation}
To achieve a high level of semantic understanding of the environment, we used gpt-4-vision-preview, the latest model available at the time of our experiments, which processes RGB images captured by the robot's front and rear cameras.
Before processing, the images are resized to $640 \times 427$ pixels. The inputs to the \ac{VLM} are the images and a text prompt that contains several components: a task description, details about the robot's capabilities, guidelines for generating each output field, the desired output format, and previously generated memory outputs and estimates for context.

The prompt is divided into several sections. The ``Task'' section provides the general instructions, explaining that the VLM should analyze the image to inform the robot's navigation. The ``Robot'' section describes the robot's capabilities as a quadruped optimized for terrain navigation. It explains that the robot can handle uneven surfaces, including stairs and gravel, but cannot cross water or jump over large obstacles. The ``Guidelines'' section specifies the required output constraints. For example, the model must avoid using directional labels such as ``left,'' ``right,'' or ``ahead,'' as well as similar or confusing labels at the same time (such as ``grass'' and ``grassy\_area''). These measures were implemented after observing that unconstrained output sometimes resulted in undesirable or overly complex semantic labels (such as ``left\_road'' and ``sunny\_sky''). The guidelines ensure that the model produces labels that can be directly segmented by the open-vocabulary semantic segmentation system without ambiguity.

We define specific output fields to capture different aspects of semantic understanding. The risk field assigns each label a numerical risk value ranging from 0.0 (safe) to 1.0 (risky). High-risk categories include trees, people, cars, fences, walls, and buildings. The steppable field indicates whether a surface can be stepped on safely (0.0 for non-steppable to 1.0 for fully steppable), regardless of its risk level. The preferred area field rates the preference of the terrain for navigation from 0.0 (not suitable) to 1.0 (highly suitable). For example, paths and trails receive higher values than grass, gravel, rocks, or bushes.

A memory mechanism preserves the context over time. After each analysis, the VLM outputs a memory text summarizing the robot's current situation. The last three memory outputs are added to the prompt. In addition, the outputs of the last three label values are averaged to stabilize labeling decisions and serve as contextual input for subsequent queries. 
At the end of the prompt, we provide a structured JSON format as the output.
The full prompt is shown in fig.~\ref{fig:full-prompt}.

After obtaining semantic labels and associated field values from the VLM, we employ an open-vocabulary semantic segmentation network based on the pre-trained model of Xu et al.~\cite{xu2023side}. This network takes the labels from the VLM analysis, augmented with an ``other'' category, and produces per-pixel probabilities for each label. These probabilities are combined with the label-specific values using the following formula:

\begin{equation}
\text{field\_value} = \sum_{\mathrm{label}} v_{\mathrm{label}} \cdot p_{\mathrm{label}}
\end{equation}
Here, $v_{\mathrm{label}}$ is the value assigned to each label (for a given field), and $p_{\mathrm{label}}$ is the probability of that label for a pixel. The probability $p_{\mathrm{label}}$ is calculated using a temperature-scaled softmax function:

\begin{equation}
p_{\mathrm{label}} = \frac{\exp(z_{\mathrm{label}} / \tau)}{\sum_{i} \exp(z_{i} / \tau)}
\end{equation}
\noindent where $z_{\mathrm{label}}$ is the logit output from the segmentation network for the label, $\tau$ is the temperature parameter ($\tau > 0$), and $i$ iterates over all labels.
For example, if the risk values are defined as path (0.1), grass (0.5), and lake (1.0), the risk value for each pixel is calculated as:
\begin{equation}
\text{risk\_value} = 0.1 \cdot p_{\mathrm{path}} + 0.5 \cdot p_{\mathrm{grass}} + 1.0 \cdot p_{\mathrm{lake}} + 1.0 \cdot p_{\mathrm{other}},
\end{equation}
where $p_{\mathrm{other}}$ is the probability of the appended ``other'' label and we also define the value for this ``other'' label per output field. We used 1.0 for risk and steppable and 0.0 for preferred\_area in our experiments.

We queried the VLM every 10 seconds to obtain new labels and values, whereas the semantic segmentation ran continuously at 6~Hz. Since the environment often changes slowly, this approach provides a stable, near-real-time semantic understanding of the scene. The segmentation step, which is much faster than the VLM queries, ensures that the elevation mapping module receives frequent updates.

\subsubsection*{Sign detection}
Although our pipeline maintains semantic label updates every 10 seconds, important transient information---such as the presence of signs---may be missed in this interval. To address this, we run a YOLO detector~\cite{yolov8_ultralytics} in parallel to detect signs in the robot's field of view. When a sign is detected, we pause the robot and append the sign's region into the VLM image query, ensuring immediate attention to the sign's content. This approach improves the detection and interpretation of critical signs, even with a high-field-of-view camera (\SI{198}{\degree}), which can make the sign appear small. This allows the robot to correctly interpret the sign and update its navigation strategy accordingly.

\subsubsection*{Semantic information integration with elevation mapping}
To incorporate semantic information extracted from RGB images into the robot's navigation and locomotion processes, we integrate this data into a 2.5D elevation map. This map fuses geometric measurements obtained from depth cameras and LiDAR point clouds with the semantic fields---such as risk, steppable, and preferred\_area---provided by the VLM pipeline. We adopt the framework developed by Miki et al.~\cite{mikielevation2022} and Erni et al.~\cite{erni2023mem}, which is a mapping package that runs on the robot's GPU.

By assigning each semantic field to its own layer, the elevation map maintains a multi-dimensional, temporally consistent representation of the environment. This approach effectively preserves past data, as opposed to relying solely on the most recent image input. As a result, the elevation map supports both local navigation and locomotion planning: the navigation module uses the elevation, risk, and preferred\_area layers to make geometrically and semantically informed path decisions, whereas the locomotion module relies on the elevation and steppable layers to determine optimal foot placement. In our experiments, we used a \SI{4}{\centi\meter} resolution over an \SI[parse-numbers = false]{8\times8}{\meter} area to maintain enough detail and context for both local navigation and locomotion.

\subsection{Locomotion control}
Our locomotion control framework builds on the work of Miki et al.~\cite{miki2022wild, miki2024confined}, extending it with additional observational inputs derived from the \ac{VLM}. In simulation, we generate steppable and non-steppable regions using Perlin noise patterns. Specifically, we randomly sample a threshold and assign a value of 1.0 to areas above this threshold, and 0.0 to areas below it. The corresponding steppable information is sampled in a circular pattern identical to the height scans and then concatenated with the height data. This combined exteroceptive observation is provided to the neural network policy.

A penalty is applied whenever a foot makes contact with a non-steppable region, guiding the policy to avoid stepping on the hazardous areas while following the commanded velocity. More concretely, we assign a negative reward of \(-\alpha\) when a foot steps onto a non-steppable region, where \(\alpha\) is a scaling coefficient that determines the penalty's relative importance within the overall reward scheme. We set \(\alpha = 0.1\) in our implementation, providing a moderate but effective incentive that encourages the policy to maintain stable and efficient locomotion while avoiding harmful foot placements.

To achieve a robust control policy suitable for real-world deployment, we employ a teacher-student reinforcement learning framework~\cite{chen2020learning,lee2020learning,miki2022wild,miki2024confined}. The training process consists of two stages.

In the first stage, a teacher policy is trained in simulation with privileged information not available on the real robot. This includes perfect exteroception and friction coefficients per foot. The policy receives height and steppable scans as exteroceptive input, along with proprioceptive data such as velocity commands, base height, roll and pitch commands, base linear and angular velocities, base orientation (gravity vector) and joint velocities and positions over time. The reward function follows the structure of previous work~\cite{miki2022wild,miki2024confined}, combining velocity-tracking rewards and penalties on joint velocities and torques, augmented by the additional penalty for stepping on non-steppable regions.

In the second stage, the trained teacher policy is distilled into a student policy. We introduce a larger observation noise into the exteroceptive inputs to make it robust in real world conditions, following the noise model described in~\cite{miki2022wild}. However, we do not add large noise to the steppable scan inputs, ensuring that the policy remains sensitive to this crucial information and does not learn to ignore it.

During deployment, the student policy receives exteroceptive inputs from an elevation map, as described in the previous section. By incorporating elevation and steppable layers, the policy can make informed decisions about foot placement in real-time. The use of steppable abstraction enables us to train the locomotion policy in simulation without the need for photorealistic rendering, and still benefit from the high-level semantic understanding provided by vision-language models on the robot.

\subsection{Local navigation policy}
Our local navigation control method builds upon the RL-based approach by Lee et al.~\cite{lee2024learning}, extending it to integrate both semantic segmentation and geometric measurements. This module generates navigational commands (base velocity and yaw rate) at 10~Hz.

The local navigation policy ensures efficient and safe navigation by leveraging geometric and semantic information.
The policy also continuously adjusts its path based on real-time feedback from the environment, enabling robust and reliable performance even in dynamic and unpredictable conditions.

\subsubsection*{Incorporating semantic information}
Building on the terrain elevation map input used by Lee et al.~\cite{lee2024learning}, we incorporate additional semantic information, specifically risk and preferred\_area, obtained from the VLM module.
The risk information identifies areas that the robot must avoid, whereas the preferred\_area highlights favorable terrain for navigation.

 For the risk information, an additional risk map is introduced alongside the elevation map as input to the policy. During training, high penalties are applied to paths that cross high-risk areas, discouraging the robot from traversing them.

For the preferred area, we extend the human-avoidance strategy proposed by Lee et al.~\cite{lee2024learning}, which introduced a virtual height increment for human-detected areas to discourage traversal. We generalize this idea by applying a virtual height increment to any unpreferred areas detected by the VLMs. Specifically, when the robot is within a preferred area, the height of surrounding unpreferred areas is artificially increased, encouraging the robot to remain on the preferred path. For instance, if the robot is navigating a preferred walking path, the adjacent grass areas are assigned a higher virtual height, making them less appealing for traversal. Conversely, when the robot is already in a non-preferred area, no height increment is applied, ensuring the robot can still navigate effectively from its current location.

\subsubsection*{Policy training}
The local navigation policy is trained in a simulated environment similar to Lee et al.~\cite{lee2024learning}, with a traversability graph generated following probabilistic roadmap--based RL (PRM-RL)~\cite{faust2018prm}. Risky areas, defined as unapproachable regions, are identified by simulating robots with a pre-trained locomotion policy. Each node in the navigation graph is labeled as either traversable or not, and a negative reward of $-1.0$ is assigned when the closest node to the robot's foot is marked as untraversable.

During training, the robot starts on a randomly selected obstacle-free node and follows a predefined sequence of nodes. The path (sequence of nodes) is replaced with a global plan during deployment. Navigation goals are determined using the pure pursuit algorithm, selecting the next waypoint from the locally transformed global path based on a fixed look-ahead distance of 8 meters. This ensures the robot consistently moves toward reachable, path-aligned goals in the environment.

\subsection{Global path planning}
Although the local navigation policy enables the robot to walk to a given waypoint within its line of sight, following a long path beyond the robot's field of perception requires global information. To address this, we implemented a global planner that outputs waypoints for the local navigation policy.

The global planner utilizes satellite data in the WGS84 coordinate system, also known as the longitude-latitude-altitude (LLA) coordinate system. The user selects a map section via the corner coordinates, which is downloaded using the Web Map Tiling Services of the Swiss Geo Portal~\cite{swisstiling}. The satellite image tiles, provided in the Cartesian LV03 coordinate system---a locally used Swiss reference frame---are transformed to LLA using the approximation given by swisstopo~\cite{swisstransform}.

For global path planning, we developed a GUI that displays the downloaded geo-referenced satellite image. The robot operator can set global goal points with arbitrary distances by clicking on the map, delete points, and zoom in for easy operation. To plan a path between the goal points, we created a server based on OpenRouteService, which incorporates Swiss hiking paths. The server takes the goal points as input and outputs sequential waypoints in LLA coordinates in \SIrange{2}{5}{\meter} intervals. In order to enable the robot to follow these global waypoints, they are continuously transformed into the local robot frame.

\subsection{Global and local pose estimation through holistic fusion}
Accurate and precise global and local pose estimation allows the robot to follow long-range and large-scale goals reliably. To enable such an estimation, the optimization-based sensor fusion framework Graph-MSF~\cite{gmsf, nubert2025holistic} is developed further to overcome the particular (practical) challenges of long-range outdoor hiking, namely large distances potentially leading to numerical issues, unreliable \ac{GNSS} reception, drifting non-global measurements, and environments poor in geometrical features.
To provide accurate and reliable robot poses as part of a map-based SLAM framework, particularly in cases when the \ac{GNSS} signal is lost, an improved version of the degeneracy-aware X-ICP~\cite{xicp, tuna2025informed, jelavic2022open3d} \ac{LiDAR}-based SLAM framework is used. 
Graph-MSF then fuses \ac{IMU} measurements together with the globally accurate but noisy \ac{GNSS} measurements, the more consistent but drifting \ac{LiDAR}-SLAM $\mathrm{SE}(3)$-pose measurements, and locally consistent TSIF~\cite{tsif}-based inertial-kinematic odometry measurements. 
Graph-MSF uses a factor graph formulation, where the state variables of the robot and the relative context are estimated as the \ac{MAP} estimate
\begin{equation}
        \states^\star = \argmax_{\states} p(\states|\Measurements) \propto p(\Measurements|  \states)p(\states).
\end{equation}
Here, $\states$ denotes the full set of robot states and dynamic state variables $\{ ^I \states_{N_I}, ^G \states_{N_G}, ^R \states_{N_R} \}$, with $^I \states_{N_I}$ denoting the robot state in the world frame, whereas $^G \states_{N_G}$ and $^R \states_{N_R}$ denote global- and reference-frame alignment states.
The direct fusion of absolute \ac{LiDAR} poses into the estimation and the automatic alignment of the map and world frame as part of the optimization are of particular importance, in contrast to treating them as local odometry measurements as usually done in earlier work~\cite{gmsf,zhao2021super,wisth2022vilens}. 
The corresponding alignment state in $^R \states_{N_R}$ at time $k$ between the world frame $\mathtt{W}$ and the \ac{LiDAR} map frame $\mathtt{M}_{\text{o3d}}$ is denoted as $T_{\mathtt{W} \mathtt{M}_{\text{o3d}}}$.
This allows the fusion framework to achieve smoother estimates, determine the robot's state in the map frame at any point, and correctly model the covariance growth based on the assumed drift of the \ac{LiDAR} mapping framework.

\subsection{Statistical analysis}
Field deployments in Figs.~\ref{fig:result_zermatt} to \ref{fig:arche} were performed once per method (N = 1) and are reported descriptively using distance, elevation change, operator interventions, and time-based autonomy rate. Mannequin avoidance (fig.~\ref{fig:mannequin}) used N = 10 trials per condition. For VLM output abstractions (Fig.~\ref{fig:vlm}C and Table~\ref{tab:vlm-comparison}; N = 27 common scenes), safety and preferred ratios were computed over pooled trajectory pixels, with 95\% bootstrap confidence intervals (10,000 resamples) and two-sided Wilcoxon signed-rank tests against ROAM A* with Holm correction across the 16 comparisons. Localization metrics (table~\ref{table:loc_quantitative}) report mean and SD, VLM benchmark scores (table~\ref{tab:vlm_results}) are micro-averaged over 372 TartanGround and 12 sign images, and prompt-ablation scores (tables~\ref{tab:prompt_ablation} and \ref{tab:sign-comparison}) are macro-averaged across the same 372 TartanGround and 12 sign images. Underlying values are provided in Data S1.

\noindent\textbf{Movie 1. ROAM navigating autonomously in the wild.} The movie introduces the ROAM pipeline and shows deployments in diverse indoor, urban, and natural environments. It then shows the controlled experiments on the H\"onggerberg campus in Zurich, in which the robot avoids a mannequin lying on the path and follows construction signs, and the hike in Zermatt, Switzerland. The movie is available at \url{https://youtu.be/xcqVpbgklD4}.

\noindent\textbf{Movie S1. Autonomous hike around Seealpsee.} The full 2.4-km hike around the lake, filmed from behind the robot. Available at \href{https://youtu.be/1TAuPEN80WE}{https://youtu.be/1TAuPEN80WE}.

\noindent\textbf{Movie S2. Reading construction signs.} The robot leaves the gravel path when a sign tells it to and returns when a second sign says so, next to a geometry-only baseline that cannot read the signs. Available at \href{https://youtu.be/LiX3yLj3ffE}{https://youtu.be/LiX3yLj3f{}fE}.

\noindent\textbf{Movie S3. System components.} The VLM risk image, the local navigation policy, and the locomotion policy, each tested on its own. Available at \href{https://youtu.be/0iFn34GOyfY}{https://youtu.be/0iFn34GOyfY}.

\noindent\textbf{Movie S4. Indoor, urban, and outdoor navigation.} The same system, without retuning, indoors, in towns, and on outdoor paths. Available at \href{https://youtu.be/wR5apHDt7WM}{https://youtu.be/wR5apHDt7WM}.

\noindent\textbf{Movie S5. Full 10.1-km autonomous hike near Zermatt.} The whole hike from Gornergrat station down to the cable car station, filmed from behind the robot. Available at \href{https://youtu.be/qaKoBthsapA}{https://youtu.be/qaKoBthsapA}.

\rpbackmatter
\section*{Acknowledgments}
We thank Yuntao Ma, Yanqing Shen, and Lukas Spirig for assistance with the Zermatt Experiment; Ryo Watanabe, Kento Kawaharazuka, Johann Schwabe, and Laurin Schmid for the Seealpsee Experiment; Yuni Fuchioka, Chenhao Li, and Yuxuan Xie for support during the development experiments; Ryo Watanabe, Cyrill P\"untener, and Johann Schwabe for software support; Fabian Tischhauser for hardware support; Tim Windecker for support with the VLM trace prediction evaluations; Jonas Frey for discussions; and Ruyi Zhou for advice on figures.
\noindent Disclaimer: None of the datasets, models, or other software used in connection with this project was created, accessed or used by Apple or any of its employees.
\paragraph*{Funding:} This work was supported in part by the Swiss National Science Foundation (SNSF) as part of projects No.~227617 and 20CH-1\_229464/1, by the European Union's Horizon Europe Framework Programme under grant agreements No.~101070596 and No.~101121321, and by Apple Inc.

\noindent Any views, opinions, findings, and conclusions or recommendations expressed in this material are those of the authors and should not be interpreted as reflecting the views, policies or position, either expressed or implied, of Apple Inc.

\paragraph*{Author contributions:}
T.M. formulated the main ideas, led the development and the entire project. He developed the vision-language model pipeline, trained locomotion policies, set up local navigation training and integrated the system. R.S. co-led the project, performed system integration and testing in both simulation and on the robot, developed the deployment infrastructure, and organized and led the hiking experiments. M.P. worked on sign detection and integrated camera-related software. T.M. and M.P. worked on the VLM abstraction ablations. T.T. developed and integrated the GNSS-LiDAR fusion. J.R. implemented global planning features. J.N. developed and fine-tuned the core algorithms for GNSS-LiDAR fusion. J.L. trained local navigation policies.
T.M., R.S., M.P., T.T., and J.R. conducted the hiking experiments; T.M., R.S., and M.P. performed analysis experiments; and T.M. and M.P. conducted the baseline comparison experiment. T.M., R.S., M.P., T.T., J.R., and J.L. conducted experiments during the development phase. T.M., R.S., M.P., T.T., J.R., and J.N. processed data from the experiments. T.M. wrote the paper with input from all authors. M.M. and M.H. contributed to conceptualization and supervision of the project.
\paragraph*{Competing interests:}
There are no competing interests to declare.

\paragraph*{Data, code, and materials availability:}
All data needed to evaluate the conclusions in the paper are present in the paper or the Supplementary Materials. The data for this study (Data S1, the image-based evaluation of the VLM output abstractions, the VLM benchmarks, and the VLM answers logged during the field deployments) have been deposited in Zenodo (DOI: 10.5281/zenodo.23492871). The code of ROAM is available at \url{https://github.com/leggedrobotics/roam}. All materials associated with this study are listed in the Methods section and/or are available commercially.

\subsection*{Supplementary Materials}
\noindent
Supplementary Methods\\
Figs. S1 to S15\\
Tables S1 to S6\\
Movies S1 to S5 (online)\\
Data S1\\
Reference (85) (listed in the main reference list)

\bibliography{bibliography}

\clearpage
\rpsupplement
\renewcommand{\thefigure}{S\arabic{figure}}
\renewcommand{\thetable}{S\arabic{table}}
\renewcommand{\theequation}{S\arabic{equation}}
\setcounter{figure}{0}
\setcounter{table}{0}
\setcounter{equation}{0}

\twocolumn[{%
\begin{center}
{\sffamily\bfseries\Large Supplementary Materials for\\[4pt] \scititle\par}\vspace{8pt}

Takahiro Miki$^{1}$,
Robin Schmid$^{1}$,
Manthan Patel$^{1}$,
Turcan Tuna$^{1}$,
Julia Richter$^{1}$ \\
Julian Nubert$^{1}$,
Joonho Lee$^{1}$,
Matthias M\"uller$^{2}$ and
Marco Hutter$^{1}$ \\
\small$^{1}$ETH Zurich, Zurich, Switzerland \\
\small$^{2}$Apple, California, USA
\end{center}

\subsubsection*{This PDF file includes:}
Supplementary Methods\\
Figs. S1 to S15\\
Tables S1 to S6\\
Caption for Data S1\\
\subsubsection*{Other Supplementary Materials for this manuscript:}
Movies S1 to S5 (online)\\
Data S1\\
\vspace{1.5em}}]

\section*{Supplementary Methods}
\subsection*{Labels during experiments}
In this section, we show how the \ac{VLM} labels evolved over time during the mission. In response to different scenes and contexts, the \ac{VLM} generated a variety of labels, each accompanied by its associated value.

\subsubsection*{Zermatt}
\begin{figure*}[htbp]
\centering
\includegraphics[width=\textwidth]{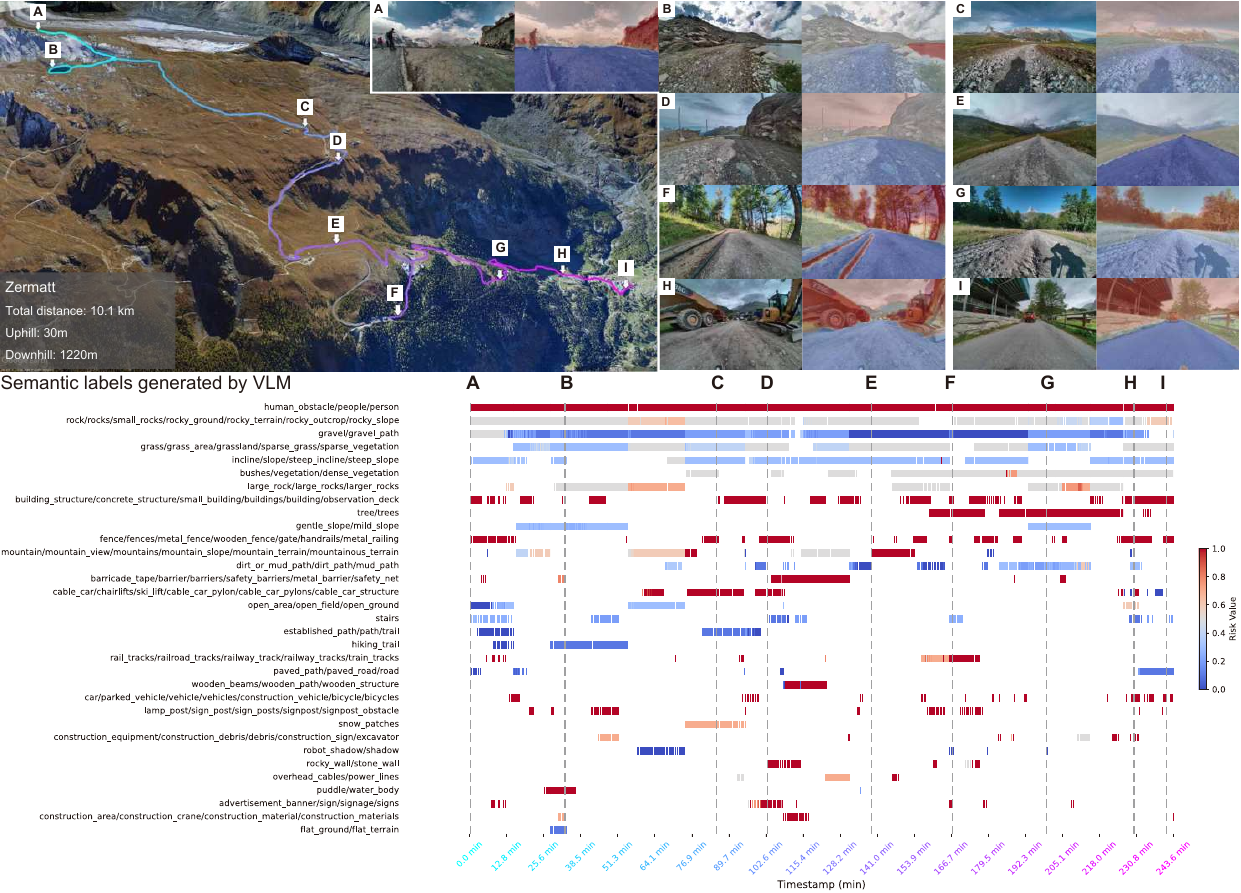}
\caption{\textbf{VLM labels during the Zermatt hike.} The labels and their risk values throughout the mission. It shows what label the VLM outputs in different scenarios as a colored bar, where the color represents the risk value and the bar represents the existence of the label.}
\label{fig:zermatt-labels}
\end{figure*}

Fig.~\ref{fig:zermatt-labels} illustrates these temporal changes, where the color indicates the assigned risk value and the bars indicate the presence of each label. For most of the mission, the gravel path was identified as a low-risk label, but this eventually shifted to labels such as paved\_path, paved\_road, and road (A to H and I). Near the lake (B), labels such as puddle and water\_body appeared. In areas where fences appeared (D), a fence or fences were detected. Similarly, when the path ran along a railroad track, labels such as rail\_track or train\_track were generated (F). To visualize the labels, we grouped similar labels into a single row. To achieve this, we utilized a \ac{VLM} to suggest appropriate categories by providing all the labels from the dataset. The categorized labels were then used to generate the figure, ensuring a clearer representation of the data.

\subsubsection*{Seealpsee}
\begin{figure*}[htbp]
\centering
\includegraphics[width=\textwidth]{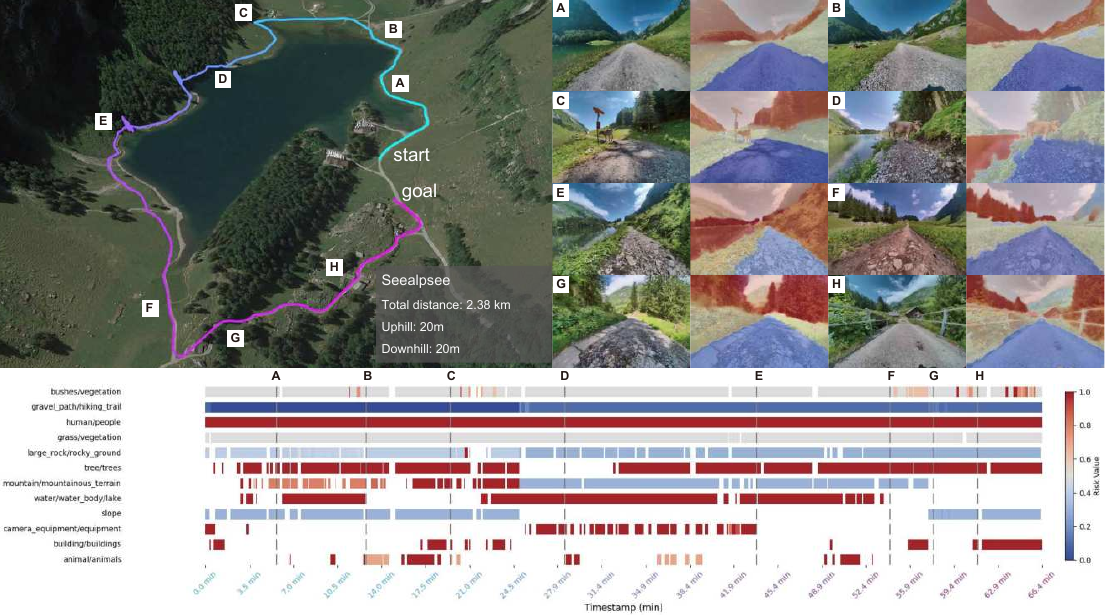}
\caption{\textbf{VLM labels during the Seealpsee hike.} The labels and their risk values throughout the mission. It shows what label the VLM outputs in different scenarios, shown as a colored bar, where the color represents the risk value and the bar represents the presence of the label.}
\label{fig:seealpsee-labels}
\end{figure*}

Fig.~\ref{fig:seealpsee-labels} also illustrates the labels during the Seealpsee mission. Throughout the majority of the mission, the traversable path was consistently identified as gravel\_path/hiking\_trail with a low-risk value (blue). This path was typically bordered by mid to high risk environmental features such as grass/vegetation or tree/trees. Where the path ran alongside the water, the label water/water\_body/lake was frequently detected (A, B, D, E). Specific obstacles were also captured. For instance, at point (D), the presence of cows on the trail triggered the detection of animal/animals as a high-risk entity. Towards the end of the trajectory, as the path approached the goal, labels such as building/buildings appeared (H).

\subsection*{Navigation from indoor to outdoor environments}
\begin{figure*}[htbp]
\centering
\includegraphics[width=\textwidth,height=0.68\textheight,keepaspectratio]{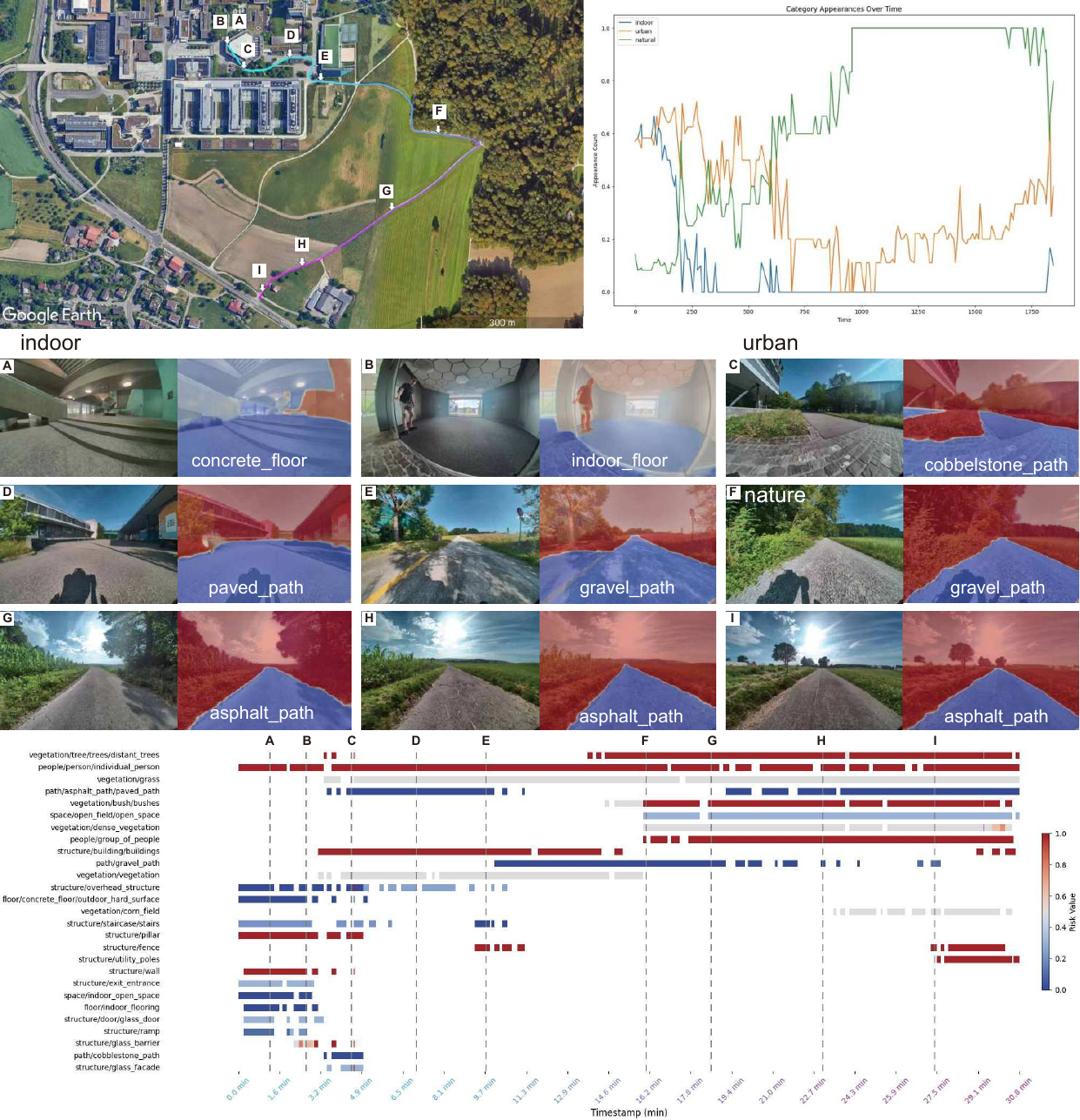}
\caption{\textbf{VLM labels during navigation from indoor to outdoor environments.} The evolution of labels and their associated risk values as the robot transitions from an indoor setting to an outdoor environment, traversing urban areas and open fields. The  category distribution plot shows the fraction of indoor-, urban-, and nature-related labels along the route.}
\label{fig:indoor-outdoor-labels}
\end{figure*}

In this experiment, we aimed to demonstrate the versatility and zero-shot adaptability of our navigation system across various environments. The experimental course was designed to encompass a broad spectrum of settings, beginning indoors and transitioning to an outdoor campus environment with paved areas and buildings, before moving through a natural area with forest and along paths bordered by grass fields. Throughout these transitions, the system exhibited adaptability in risk assessment (fig.~\ref{fig:indoor-outdoor-labels}). It seamlessly transitioned from indoor features, such as floors and stairs, to outdoor elements, including paved surfaces, grass, and buildings. As the environment shifted to natural settings, the system further adapted, recognizing gravel paths as preferable routes and trees as obstacles. Notably, the system demonstrated a nuanced understanding by dynamically adjusting risk values based on environmental variations, such as the height of grass alongside paths. Additionally, the experiment showed the system's capability to manage dynamic elements by accurately identifying and categorizing obstacles, such as bicycles or benches, as potential obstacles. These findings indicate the generalizability of our \ac{VLM}-based navigation approach. 

\subsection*{Avoiding a mannequin}
\begin{figure*}[htbp]
\centering
\includegraphics[width=\linewidth]{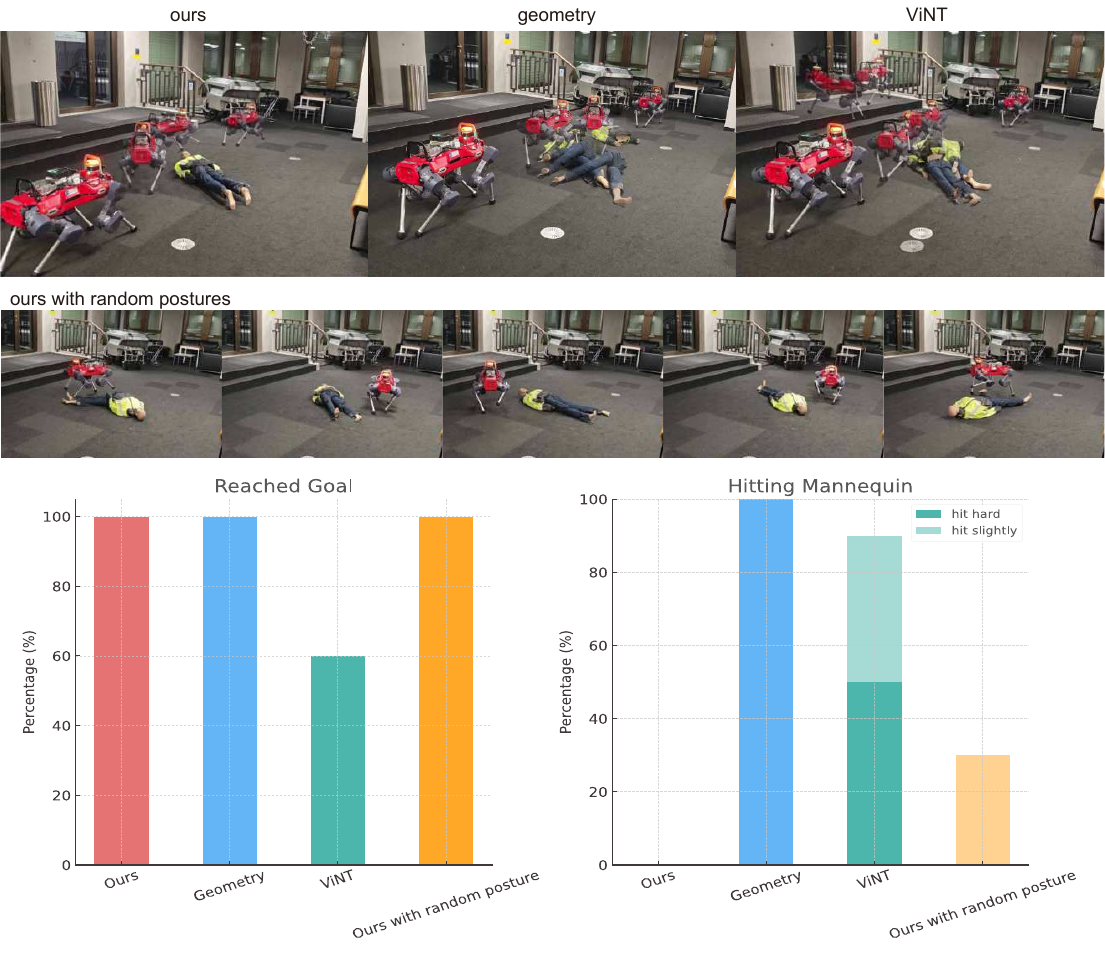}
\caption{\textbf{Mannequin avoidance.} Comparison of mannequin avoidance between our method, a geometry-based baseline, and ViNT. We placed a mannequin between the start and goal positions, ran 10 trials per condition (N = 10), and recorded both goal-reaching and collisions (classified as either a ``hard'' or a ``slight'' hit). The condition ``ours with random postures'' shows our method with the mannequin placed in random postures; collisions for the geometry baseline were not subdivided into hard and slight hits. Rates are the fraction of the 10 trials; the number of trials with each outcome is listed in Data S1.}
\label{fig:mannequin}
\end{figure*}

In this experiment, we placed a mannequin on the floor and tasked the robot with navigating to the opposite side, measuring both its ability to reach the goal and its rate of collisions with the mannequin (fig.~\ref{fig:mannequin}). We considered the trial successful if the robot arrived within \SI{2}{\meter} of the goal position. If the robot collided repeatedly or with substantial force, we labeled it as a ``hard hit,'' whereas a single, minor contact was labeled ``slight hit.'' Each experiment was repeated ten times to compute probabilities.

We first placed the mannequin in a fixed position and initialized the robot from the same location for each method. Our method was compared against a geometry-based baseline~\cite{fieldlocalplanner2022} and ViNT~\cite{shah2023vint}, the latter of which used a goal image captured without the mannequin present. Our approach reached the goal in 100\% of trials and recorded no collisions. The geometry baseline also consistently reached the goal, but collided with the mannequin in every trial. ViNT reached the goal 60\% of the time, with 90\% of runs involving collisions. The reduced success rate for ViNT may stem from the mannequin being an object not present in its goal image.

We then increased the difficulty by placing the mannequin randomly in different postures. Although our method still achieved a 100\% success rate, it incurred a 30\% ``slight hit'' rate, largely due to imprecise avoidance of thinner parts, such as the mannequin's legs. 
Overall, our method achieved higher goal-reaching performance with fewer collisions compared with the baseline approaches. The number of trials with each outcome is listed in Data S1.

\subsection*{Full prompt}
\begin{figure*}[htbp]
\centering
\includegraphics[width=\textwidth]{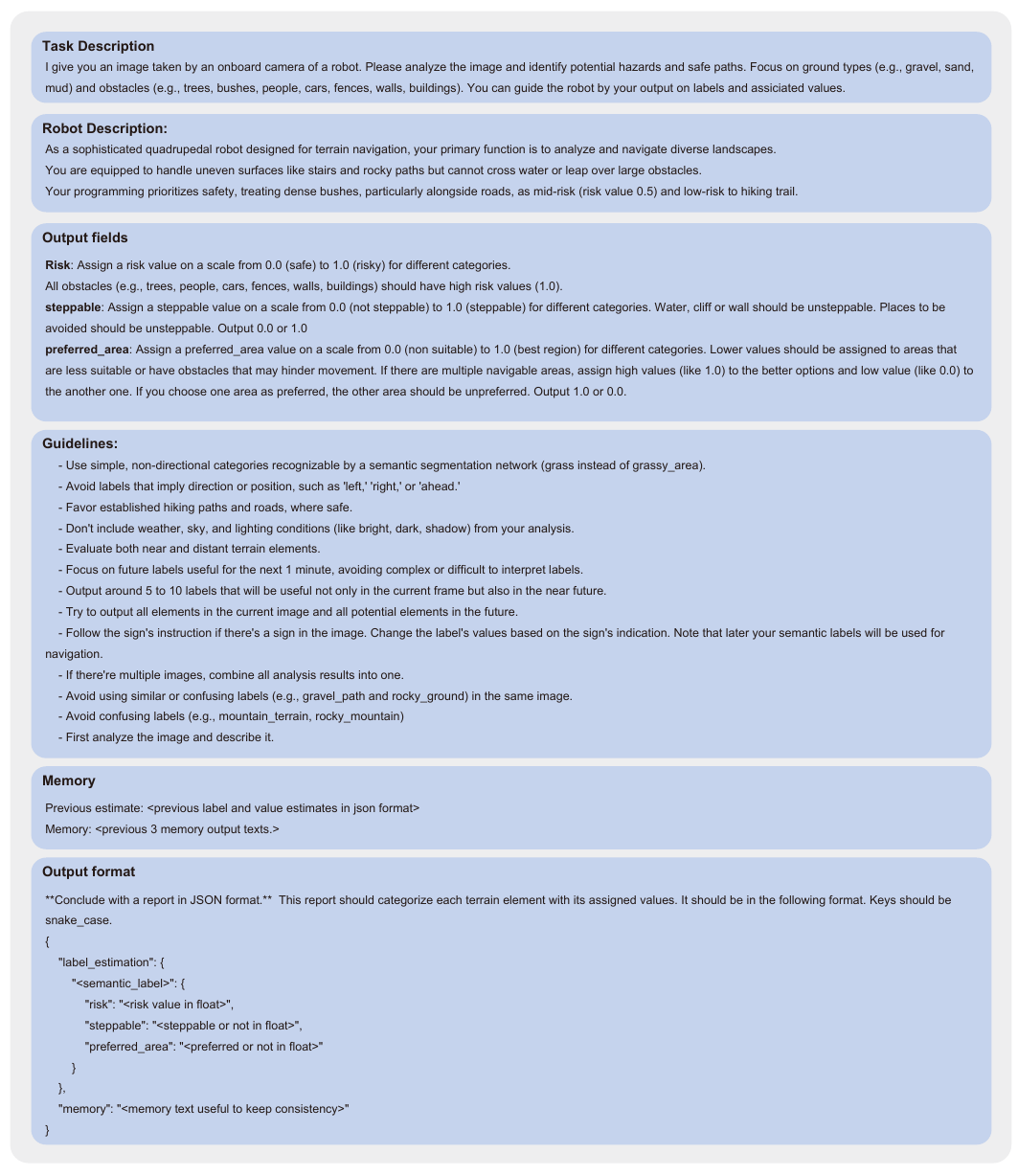}
\caption{\textbf{Full VLM prompt.} The full prompt for the VLM. We provide this prompt plus the front and rear camera images to the VLM model.}
\label{fig:full-prompt}
\end{figure*}

In this section, we present the full prompt used to query the \ac{VLM} and guide the robot's navigation decisions. The full prompt is shown in fig.~\ref{fig:full-prompt}. The prompt begins with a Task Description, where the \ac{VLM} is instructed to identify hazards and safe paths in images captured by the onboard camera, focusing on relevant terrain types (such as gravel or sand) and obstacles (such as fences or walls).
A Robot Description follows, providing key details on the quadrupedal platform's capabilities, such as navigating stairs, and its limitations, such as the inability to cross water.
The next part introduces the output fields---namely, risk, steppable, and preferred\_area---which determine hazard levels and help highlight viable paths.

The prompt then offers a set of Guidelines that ensure semantic labels remain relevant for navigation and are easily processed by the segmentation model, avoiding directional or overly complex terms. A memory mechanism incorporates the previous three outputs from the \ac{VLM} and their averaged label values, providing consistency and reducing abrupt label changes over time. Lastly, a JSON-based Output Format is specified, structuring the VLM's results into a label estimation block, which contains the assigned values for each label, and a memory field for maintaining continuity across frames. This comprehensive prompt design ensures the \ac{VLM} delivers clear, consistent, and actionable outputs to inform the robot's navigation.

\subsection*{Effect of memory in the prompt}
\begin{figure*}[htbp]
\centering
\includegraphics[width=\linewidth]{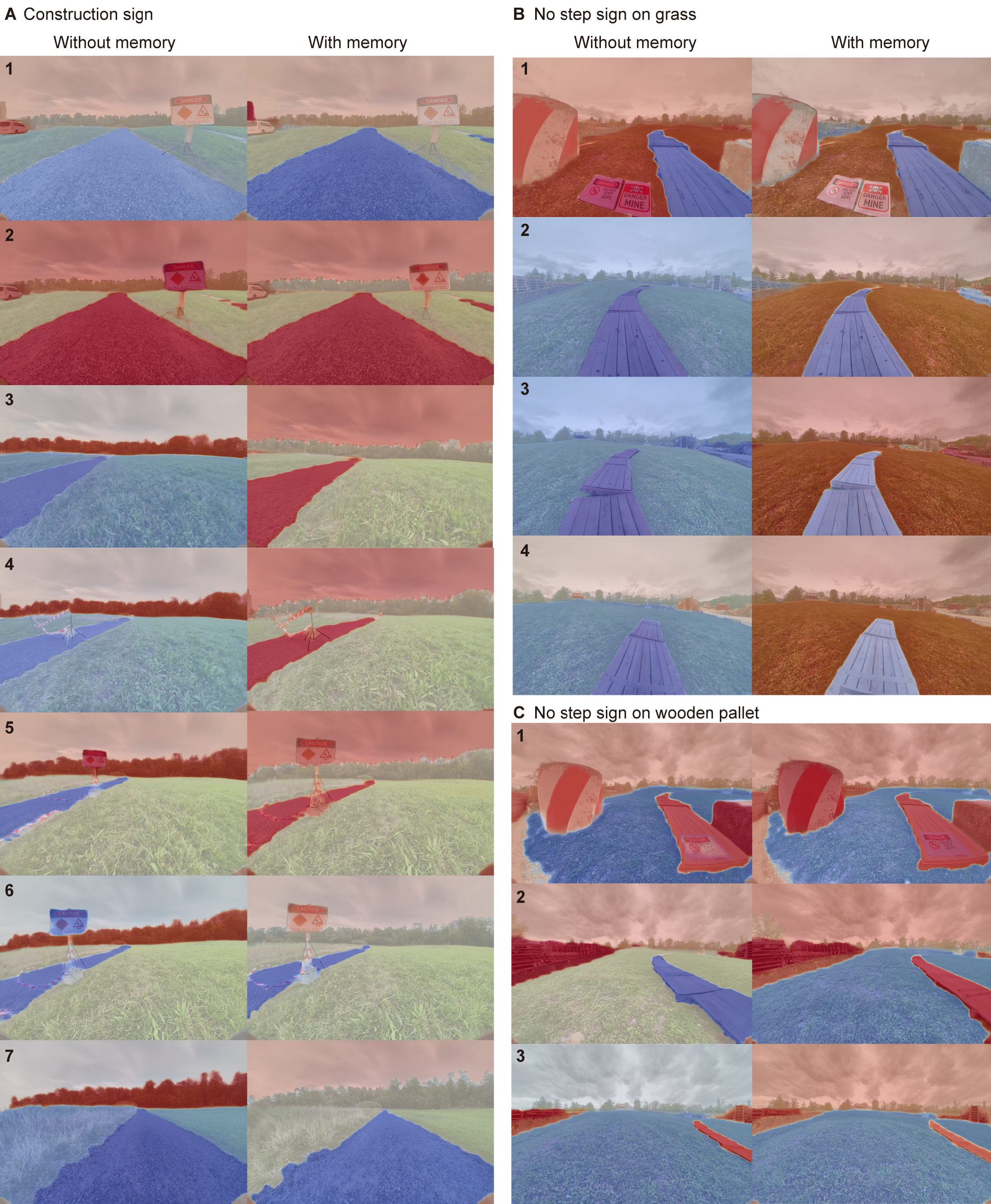}
\caption{\textbf{Effect of memory in the prompt.} Comparison of the risk with memory and without memory module in the prompt for (A) the construction sign experiment and (B and C) the no-step sign experiment.}
\label{fig:memory}
\end{figure*}

We evaluated the effectiveness of the memory module in our \ac{VLM} pipeline by using the same dataset from the construction sign and no-step sign experiments, comparing outputs generated with and without memory. As illustrated in fig.~\ref{fig:memory}A, the \ac{VLM} without memory adjusts risk values upon detecting the construction sign but quickly reverts to nominal values once the sign leaves the immediate frame, losing its context. A similar pattern appears in the no-step sign experiment in fig.~\ref{fig:memory}, B and C, where the system initially assigns correct risk values but fails to retain that information as it moves forward. In contrast, integrating memory preserves this contextual information---particularly around signs---leading to more consistent and accurate risk assignments throughout the task.

\subsection*{Limitations during the hiking experiments}
In this section, we summarize the limitations and the interventions required during our two hiking experiments at Seealpsee and Zermatt. We have categorized each of the five distinct intervention events by its dominant root cause.

At Seealpsee, there were three intervention events. The first occurred when the robot approached the lake; the shallow lake edge visually blended with the surrounding ground, and the open-vocabulary segmentation module failed to delineate it cleanly (\textbf{segmentation error}).  The operator intervened in advance to avoid potential risks. However, because the deeper part of the lake was correctly segmented and treated as high risk, this event was trajectory-dependent: if the robot's local path was planned further from the lake edge, the area could be traversed without intervention.  

The second intervention occurred on a steep, slippery slope where the path was unclear to the semantic segmentation model (\textbf{segmentation error}).  This event was compounded by the local navigation policy's limited perception horizon. Because the policy could not perceive the long-range terrain to determine whether the globally correct path lay to the left or right, it initiated a local exploration behavior. Although the policy might have eventually recovered, it chose a path sloping toward the lake, prompting the operator to preemptively intervene for safety.

Finally, the route was obstructed by a narrow, turnstile-style gate that the robot was unable to pass through due to its geometry (\textbf{physical factors}). To continue the mission, we had to briefly pause and manually carry the robot over the gate.

The Zermatt experiment required two interventions at two construction sites, both of which were highly reproducible \textbf{segmentation errors} since the specific visual features consistently confused the module. First, the main path was blocked by a fence, leaving only a narrow detour. Although the \ac{VLM} correctly labeled the fence as high risk, the segmentation model failed to detect it accurately, resulting in the robot colliding with the fence. The operator intervened to guide the robot into the detour corridor, which was so narrow that the local navigation policy occasionally stalled, requiring several short forward commands to proceed.

\begin{figure*}[htbp]
\centering
\includegraphics[width=0.8\linewidth,height=0.68\textheight,keepaspectratio]{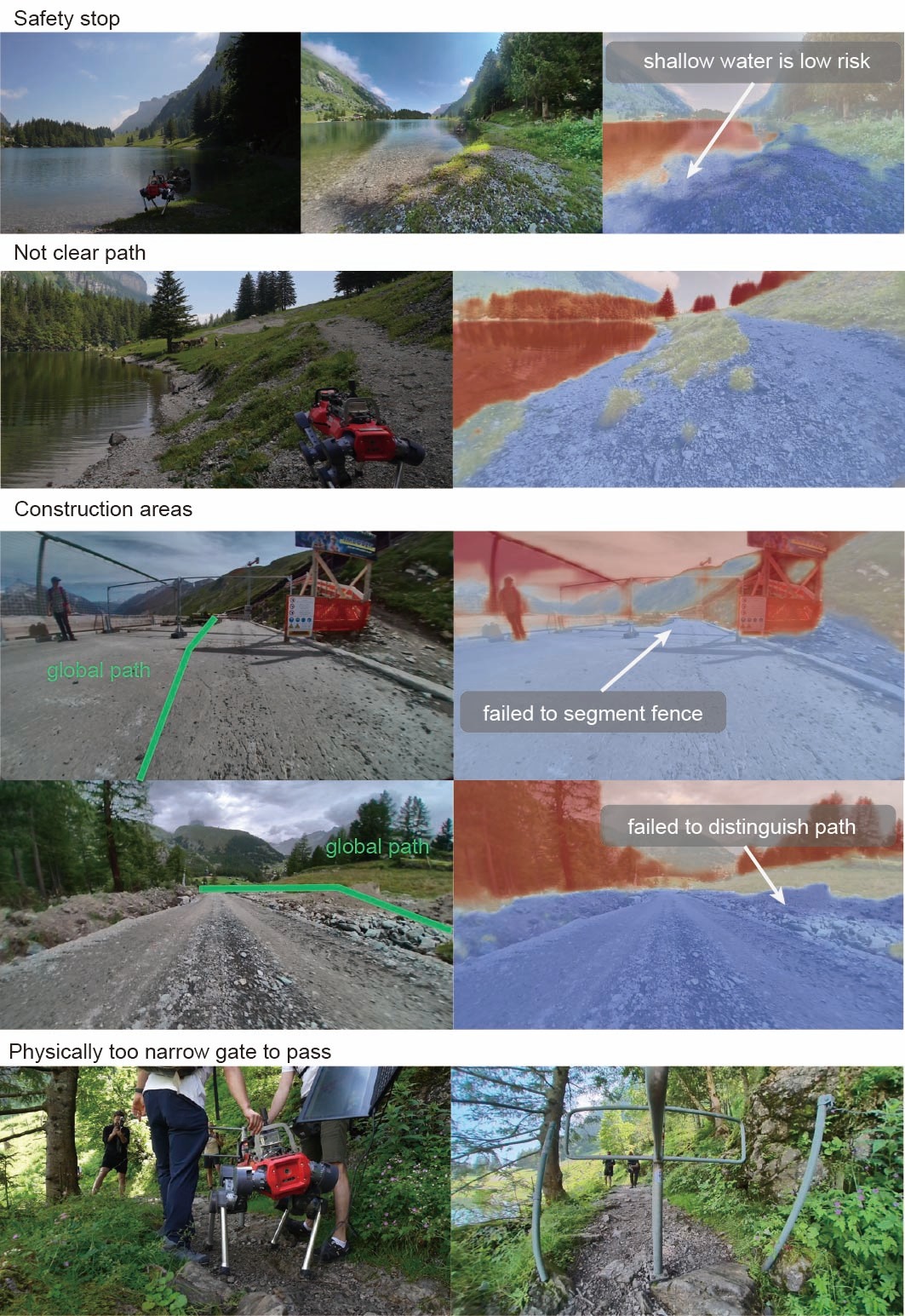}
\caption{\textbf{Interventions during the hiking experiments.} Near the lake, the operator intervened once as a precautionary measure. Another intervention was required due to a partially obscured path that impeded segmentation. Two further interventions occurred in the construction area, where fences or paths were misclassified, and the global path continued to point to the old route, causing a deviation of \SIrange{10}{20}{\meter}. Finally, an impassable gate necessitated carrying the robot over by hand.}
\label{fig:internvention}
\end{figure*}

In the second scenario, the old path indicated by the global plan was obstructed by large rocks, whereas a newly formed path was placed \SIrange{10}{20}{\meter} to the side, as shown in fig.~\ref{fig:internvention}. Although the \ac{VLM} labeled gravel\_path as a preferred area, the segmentation model failed to distinguish it from the surrounding rocky terrain, causing the robot to head toward a risky cliff. To avoid a potential fall, the operator intervened and guided the robot to a safer path.

Furthermore, although the robot mostly avoided the grass, it occasionally traversed it, since grass is considered non-preferred rather than high-risk.

In summary, of the five distinct intervention events, the dominant root cause was \textbf{segmentation errors} for 4 events and \textbf{physical factors} for 1 event. We observed \textbf{0 events} attributable to \textbf{VLM semantic errors}, \textbf{local navigation policy errors} (as the dominant cause), or \textbf{locomotion policy errors}. This breakdown indicates that the open-vocabulary segmentation step downstream of the VLM is the dominant source of failure, making it the most promising target for improvement.

\subsection*{Localization and state estimation}
We evaluate the performance of the deployed localization and state estimation solution, \ac{HF}. To successfully navigate in the environment, accurate global localization, even in the absence of reliable \ac{GNSS} reception, precise and fast local motion estimation, and consistent estimates without large jumps upon \ac{GNSS} return are necessary.

\subsubsection*{Accurate and precise state estimation in large-scale environments}

\begin{figure*}[htbp] %
    \centering
    \includegraphics[height=0.3\linewidth]{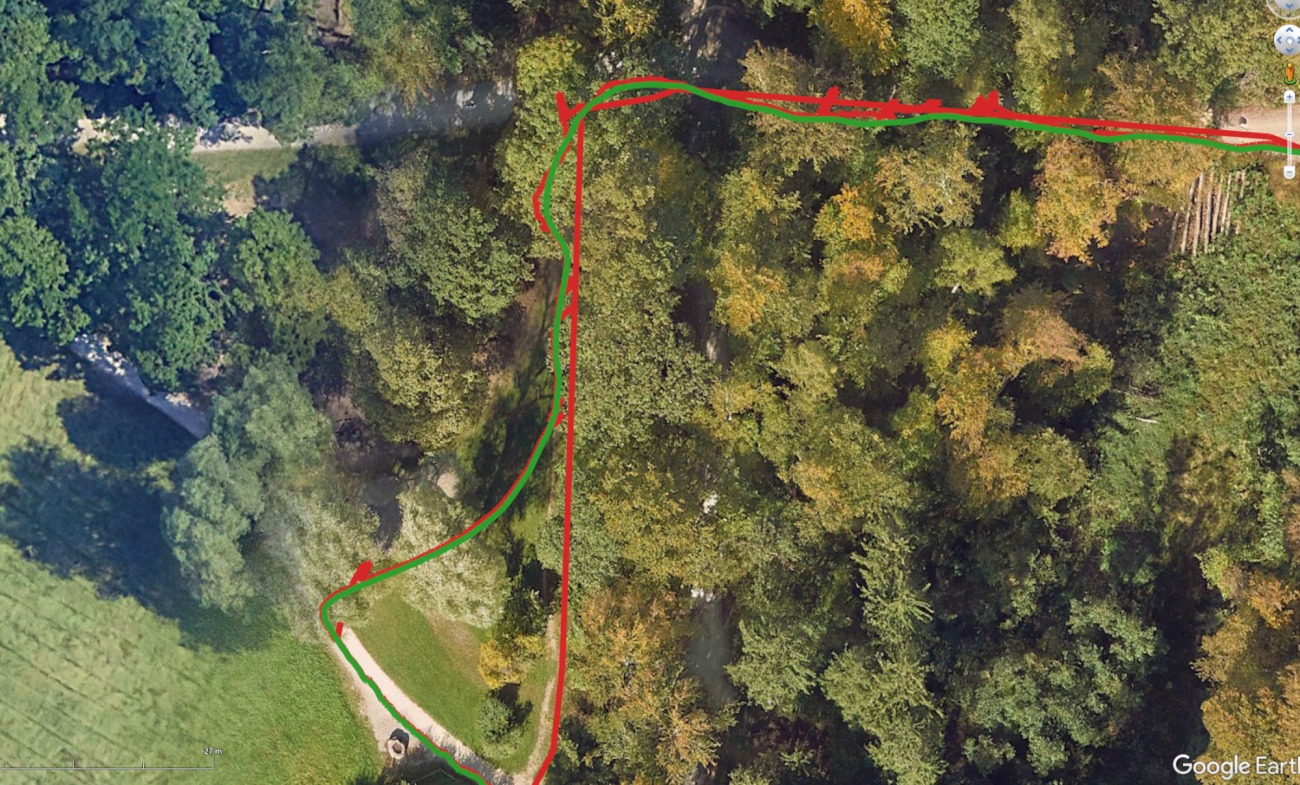}
    \includegraphics[height=0.3\linewidth]{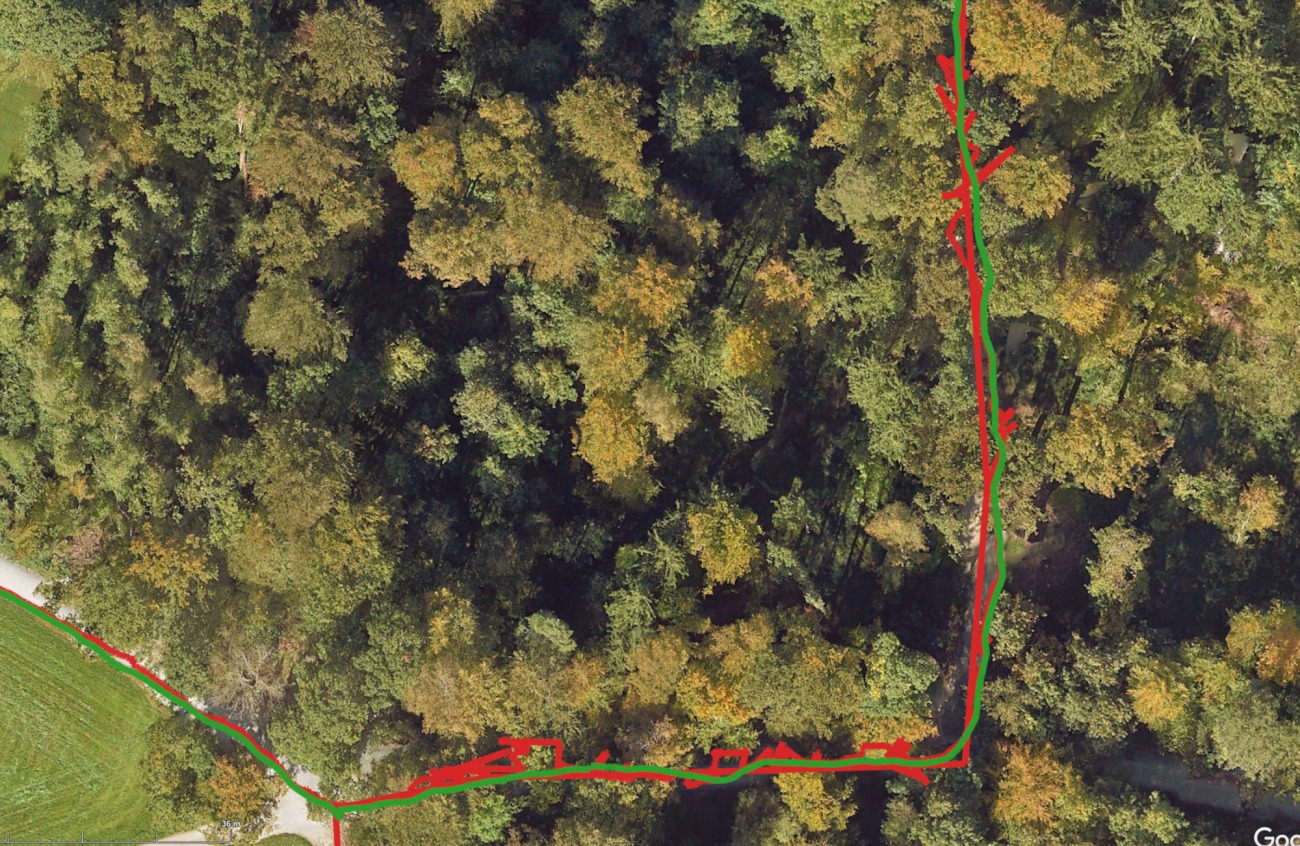}
    \caption{\textbf{Qualitative comparison of the fused and raw GNSS trajectories.} The {\color{red}red} path is the raw measured \ac{GNSS} position, whereas {\color{green}green} depicts the fused trajectory at \SI{400}{\Hz} obtained online during the operation (not replayed).}
    \label{fig:loc_hoengg_estimation_satellite}
\end{figure*}

The estimated \ac{RT} trajectory at full \ac{IMU} frequency of our solution is shown in fig.~\ref{fig:loc_hoengg_estimation_satellite}, overlaid with the raw \ac{GNSS} position measurements (for offline pseudo-\acs{GT} evaluation, see Pseudo \acs{GT} generation).
As visible in the figure, the raw \ac{GNSS} signal becomes imprecise once it enters the forest, with jumps of up to \SI{20}{\meter} occurring mostly in the $z$-direction, which challenges fusion and global alignment.
In contrast, the \ac{RT} estimate of our system combines the best of the non-drifting but imprecise \ac{GNSS} measurements and the smooth but drifting \ac{LiDAR} \ac{SLAM} poses.

\subsubsection*{Pseudo \acs{GT} generation}
\label{sec:gt_generation}
\begin{figure*}[htbp] %
    \centering
    \begin{minipage}[b]{0.48\linewidth}\centering
    \includegraphics[width=\linewidth]{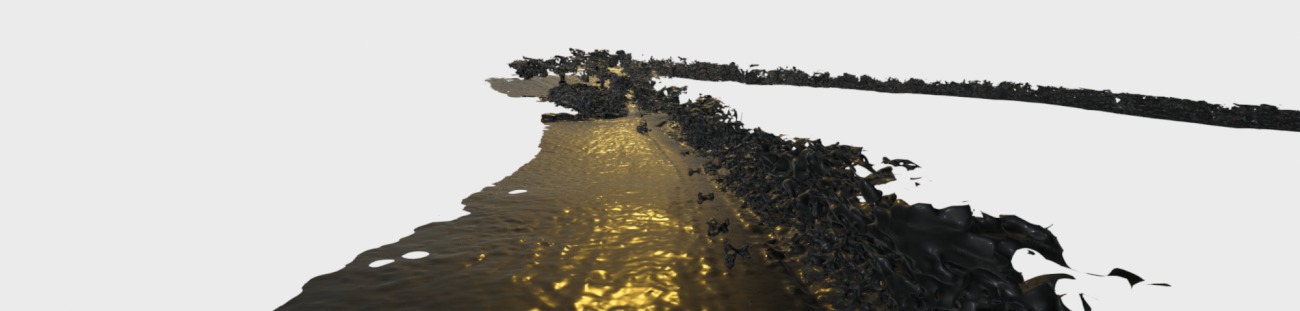}\\[2pt]
    \includegraphics[width=\linewidth]{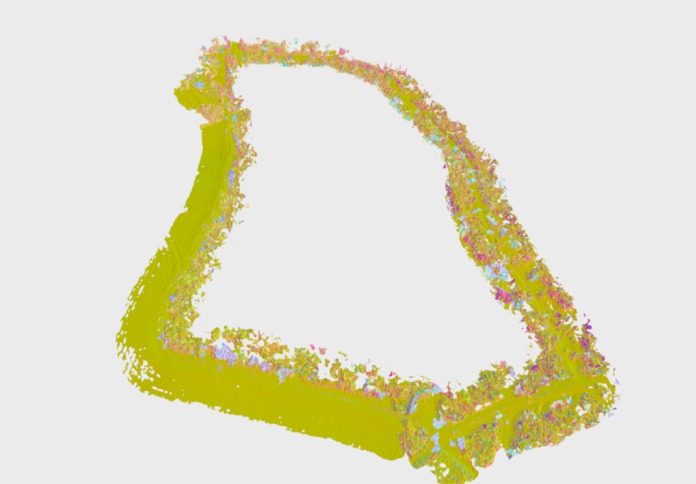}
    \end{minipage}\hfill
    \begin{minipage}[b]{0.48\linewidth}\centering
    \includegraphics[width=\linewidth]{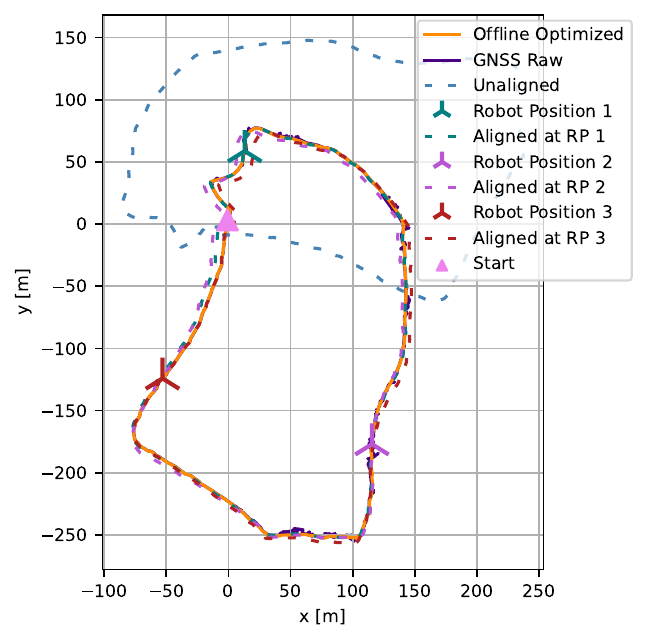}
    \end{minipage}\\
    \begin{minipage}[t]{0.48\linewidth}
    \caption{\textbf{Mesh and normal map of the generated point cloud map.}  The map is generated by accumulating the \ac{LiDAR} points in a map using the pseudo \ac{GT} estimate.}
    \label{fig:loc_offline_map_quality}
    \end{minipage}\hfill
    \begin{minipage}[t]{0.48\linewidth}
    \caption{\textbf{Dynamic alignment of the \ac{LiDAR} \ac{SLAM} trajectories by \ac{HF}.} The absolute \ac{LiDAR} measurements drift with respect to the actual robot position, which is compensated for by the automatic alignment of \ac{HF}.}
    \label{fig:loc_lidar_pose_alignment}
    \end{minipage}
\end{figure*}

To evaluate the effect of the fusion quantitatively, we run a full offline optimization over the entire trajectory. The offline-optimized trajectory is locally smooth and globally accurate, as reflected in the alignment in fig.~\ref{fig:loc_lidar_pose_alignment} and the quality of the generated mesh and normal map of the environment in fig.~\ref{fig:loc_offline_map_quality}.

\subsubsection*{Quantitative analysis}
We then compare the raw \ac{GNSS} and \ac{LiDAR} trajectories to the batch-optimized trajectory configurations against this pseudo-ground truth solution. 
Quantitative results in terms of \ac{APE}, \ac{ARE}, \ac{RPE}, and \ac{RRE} are shown in table~\ref{table:loc_quantitative}, highlighting the advantages of the fusion. 
In particular, the estimated solution is globally substantially more accurate (\ac{ARE}, \ac{APE}) than the \ac{GNSS} signal and locally substantially smoother and more consistent than the \ac{LiDAR} trajectory (\ac{RPE}, \ac{RRE}).

\begin{table*}[htbp]
\centering
\resizebox{0.75\textwidth}{!}{ 
\begin{tabular}{l|cc|cc|cc|cc}
    \hline
    \multirow{2}{*}{Method} & \multicolumn{2}{c|}{APE [\si{\meter}]} & \multicolumn{2}{c|}{ARE [\si{\degree}]} & \multicolumn{2}{c|}{RPE [\%]} & \multicolumn{2}{c}{RRE [$\frac{\si{\degree}}{\si{\kilo\meter}}$]}\\
     & $\mu$ & $\sigma$ & $\mu$ & $\sigma$ & $\mu$ & $\sigma$ & $\mu$ & $\sigma$ \\
    \hline
    \ac{LiDAR}-only      & 1.43          & 1.19          & 2.51 & 1.50 & 0.50          & 0.56 & 160.18 & 108.02 \\
    \ac{GNSS}-only       & 0.54          & 0.98          & --            & --            & 1.67          & 4.66          & --            & --   \\
    Full \acs{HF} World   & \textbf{0.44} & \textbf{0.61} & \textbf{1.08} & \textbf{1.05} & \textbf{0.46} & \textbf{0.87}          & \textbf{62.74} & \textbf{58.27} \\
    \hline
\end{tabular}
}
\caption{\textbf{Localization accuracy of the fused and single-sensor estimates.} Comparison of methods in terms of \ac{APE}, \ac{ARE}, \acs{RPE}, and \acs{RRE}. \acs{RPE} and \acs{RRE} are computed over all snippet pairs of \SI{20}{\meter} length.}
\label{table:loc_quantitative}
\end{table*}

\subsubsection*{Alignment of the drifting LiDAR trajectory}
Moreover, the deployed optimization-based factor graph solution is not only able to fuse the global \ac{GNSS} measurements with other local measurements (such as \ac{LiDAR} between factors) as usually done in literature~\cite{zhao2021super, wisth2022vilens} but can align the full path obtained through Open3D-SLAM \ac{LiDAR} map registration~\cite{jelavic2022open3d, xicp} along the entire trajectory around the current robot pose belief despite occurring drift.
This can be seen in fig.~\ref{fig:loc_lidar_pose_alignment}, where the aligned and non-aligned \ac{LiDAR} trajectories are overlaid with the pseudo-\ac{GT} trajectory at three selected robot positions (beginning, middle, end) along the mission.
It can be seen that locally the shape of the \ac{LiDAR} trajectory adequately matches the \ac{GT} trajectory due to the active alignment of \ac{HF}.

\subsection*{Evaluating different \acp{VLM}}
The primary ROAM system architecture relies on a large, cloud-based \ac{VLM} (\texttt{gpt-4-vision-preview}, as described in Materials and Methods), which, despite its powerful reasoning, introduces practical considerations for deployment in the wild. A dependency on stable internet connectivity and potential network latency can be limitations in remote and unstructured environments. This motivated an evaluation to investigate whether more compact or newer VLM alternatives could provide the necessary common-sense and contextual reasoning for autonomous navigation. To this end, we conducted a quantitative evaluation to assess the ability of a range of models to perceive and assign risk.

For this evaluation, we created a benchmark comprising two distinct challenges. The first, the TartanGround Dataset~\cite{patel2025tartanground}, utilizes 372 simulated scenes to evaluate a model's foundational ability to identify common terrain. The second, the Sign Dataset, uses 12 real-world images featuring instructional signs to assess higher-level, contextual reasoning. We benchmarked several models, including \texttt{gpt-4o}, various \texttt{gemini} and \texttt{gemma} models, and \texttt{qwen2.5-vl}, measuring performance by calculating the Micro Intersection over Union (IoU) between their generated risk maps and our ground-truth labels.

Due to the rapid iteration and deprecation cycles of commercial foundation models over the course of this research and its subsequent review period, different model versions were utilized across our evaluations. The original hardware field deployments were conducted using gpt-4-vision-preview. Following its deprecation by the provider, our comprehensive VLM benchmarking (detailed in table~\ref{tab:vlm_results}) evaluated its direct successor, gpt-4o, alongside gemini-2.0-flash and other open weight models. Most recently, as these subsequent models were also deprecated, our prompt ablation and robustness studies were conducted using the currently available gemini-3.5-flash. This necessary progression across multiple model generations empirically demonstrates the model-agnostic nature of the ROAM architecture, confirming that the system can seamlessly integrate with and benefit from advancing foundation models without requiring structural modifications to the underlying control policies.

\begin{table*}[htbp]
\caption{\textbf{Performance of different VLMs on the TartanGround and Sign benchmarks.} IoU, precision, and recall are micro-averaged over all pixels of the 372 TartanGround images and the 12 sign images.}
\centering
\resizebox{0.75\textwidth}{!}{%
\begin{tabular}{l|ccc|ccc}
\hline
\multirow{2}{*}{Model} & \multicolumn{3}{c|}{TartanGround~\cite{patel2025tartanground}} & \multicolumn{3}{c}{Sign} \\
      & IoU & Prec. & Rec. & IoU & Prec. & Rec. \\
\hline
gemini-2.0-flash         & 0.851 & \textbf{0.901} & 0.938 & \textbf{0.912} & \textbf{0.944} & \textbf{0.964} \\
gemini-2.0-flash-lite    & 0.839 & 0.882 & 0.945 & 0.799 & 0.934 & 0.846 \\
gpt-4o                   & \textbf{0.853} & 0.895 & \textbf{0.947} & 0.790 & 0.863 & 0.904 \\
gemma-3-27b              & 0.836 & 0.887 & 0.936 & 0.686 & 0.795 & 0.833 \\
gemma-3-4b               & 0.837 & 0.886 & 0.939 & 0.521 & 0.697 & 0.674 \\
qwen-2.5-vl-32b          & 0.800 & 0.880 & 0.897 & 0.562 & 0.863 & 0.617 \\
qwen-2.5-vl-7b           & 0.656 & 0.836 & 0.753 & 0.416 & 0.710 & 0.501 \\
\hline
\end{tabular}
}
\label{tab:vlm_results}
\end{table*}

The results, shown in table~\ref{tab:vlm_results}, reveal a distinction in model capabilities. On the TartanGround dataset, most models performed well, demonstrating a solid grasp of basic scene understanding. Top contenders like \texttt{gpt-4o} and \texttt{gemini-2.0-flash} achieved high IoU scores of 0.853 and 0.851, respectively, and the smaller \texttt{gemma-3-27b} model scored a strong 0.836. This suggests that the ability to identify standard environmental features is well-integrated across many modern VLMs.
However, on the more challenging Sign Dataset, which requires associating a textual command with a physical object, a bigger performance gap emerged. Here, \texttt{gemini-2.0-flash} excelled with an IoU of 0.912, correctly interpreting the signs and their context. In contrast, the performance of smaller models dropped sharply, with \texttt{gemma-3-4b} and \texttt{qwen-2.5-vl-7b} dropping to 0.521 and 0.416. The reason for this failure became apparent upon qualitative inspection: although smaller models could often read the text of a sign, such as ``DO NOT STEP,'' they failed to semantically ground this instruction to the correct object in the scene, like the wooden pallet it was placed on. This failure to bridge language with spatial context is a major shortcoming.

\subsection*{Prompt ablation and robustness studies}
\label{sec:prompt_ablation_details}

To understand the impact of our prompt engineering choices on the semantic risk mapping accuracy, we perform a systematic prompt ablation and perturbation study. We evaluate these variations across both the Sign dataset and the TartanGround dataset.
We used \texttt{gemini-3.5-flash} for this analysis because \texttt{gemini-2.0-flash} and \texttt{gpt-4o} had been deprecated when this ablation study was added.

\subsubsection*{Ablation configurations}

Each ablation is designed to evaluate a distinct prompt component in the overall system instructions. We compare five prompt ablations and two fixed-label baselines against the baseline (full prompts) system.

No robot description. In the baseline configuration, the \ac{VLM} is given a detailed physical description of the robotic platform (here, an ANYmal quadruped) to align risk scores with the robot's actual mobility profile:
\begin{lstlisting}
Robot description:
As a sophisticated quadrupedal robot designed for terrain navigation, your primary function is to analyze and navigate diverse landscapes. You are equipped to handle uneven surfaces like stairs and rocky paths but cannot cross water or leap over large obstacles...
\end{lstlisting}
In this ablated version, this detailed description is replaced by a generic, platform-agnostic agent prompt:
\begin{lstlisting}
Robot description:
You are an autonomous robot navigating through an environment. You prioritize safety and prefer stable, non-hazardous paths.
\end{lstlisting}

No directional ban. To ensure that the generated semantic labels represent purely spatial, non-positional classes suitable for general segmentation networks, the baseline explicitly prohibits directional modifiers:
\begin{lstlisting}
- Avoid labels that imply direction or position, such as 'left,' 'right,' or 'ahead.'
\end{lstlisting}

No terrain guidelines. The baseline contains specific heuristic instructions to handle terrain boundaries, temporal relevance, and category consolidation:
\begin{lstlisting}
- Favor established hiking paths and roads, where safe.
- Don't include weather, sky, and lighting conditions (like bright, dark, shadow) from your analysis.
- Evaluate both near and distant terrain elements.
- Focus on future labels useful for the next 1 minute, avoiding complex or difficult to interpret labels.
- Avoid using similar or confusing labels (e.g., gravel_path and rocky_ground) in the same image.
- Avoid confusing labels (e.g., mountain_terrain, rocky_mountain)
\end{lstlisting}
This ablated run evaluates the VLM's performance when these specialized navigational constraints are omitted.

No guidelines. This represents the minimal prompt configuration. We remove the entire guidelines structure, retaining only the core task instruction and the final output instruction:
\begin{lstlisting}
Guidelines for your analysis:
- First analyze the image and describe it.
\end{lstlisting}

Perturbation. To test if the VLM relies on brittle phrasing or exact string matches, we introduce this configuration. Every system key and numeric metric label in the baseline prompt is replaced by a distinct synonym or rephrased term, keeping the semantic intent identical: ``Risk'' is replaced by ``Danger Level'' (``Assign a danger level on a scale from 0.0 (secure) to 1.0 (hazardous)...''), ``Steppable'' by ``Traversability'' (``Assign a traversability value on a scale from 0.0 (impassable) to 1.0 (traversable)...''), and ``Preferred Area'' by ``Recommended Zone'' (``Assign a recommended\_zone value on a scale from 0.0 (unsuitable) to 1.0 (optimal sector)...'').

Fixed-label baselines. To demonstrate the necessity of VLM-driven dynamically generated semantic categories, we compare our system against two static lookup tables representing traditional, pre-defined semantic categorization systems. The ``Fixed Labels'' table is a comprehensive, statically compiled dictionary mapping 55 specific, fine-grained spatial classes directly to heuristic traversability values. This represents a typical, fine-grained human-designed classification table. The ``Fixed Labels (Simple)'' table is a simplified semantic categorizer containing only 9 general, broad-scope classes to simulate lower-fidelity baseline classifiers (tables~\ref{tab:fixed_labels} and \ref{tab:fixed_labels_simple}).

\subsubsection*{ROAM is robust to prompt variations but not to fixed label sets}
\begin{figure*}[htbp]
  \centering
  \includegraphics[width=\textwidth]{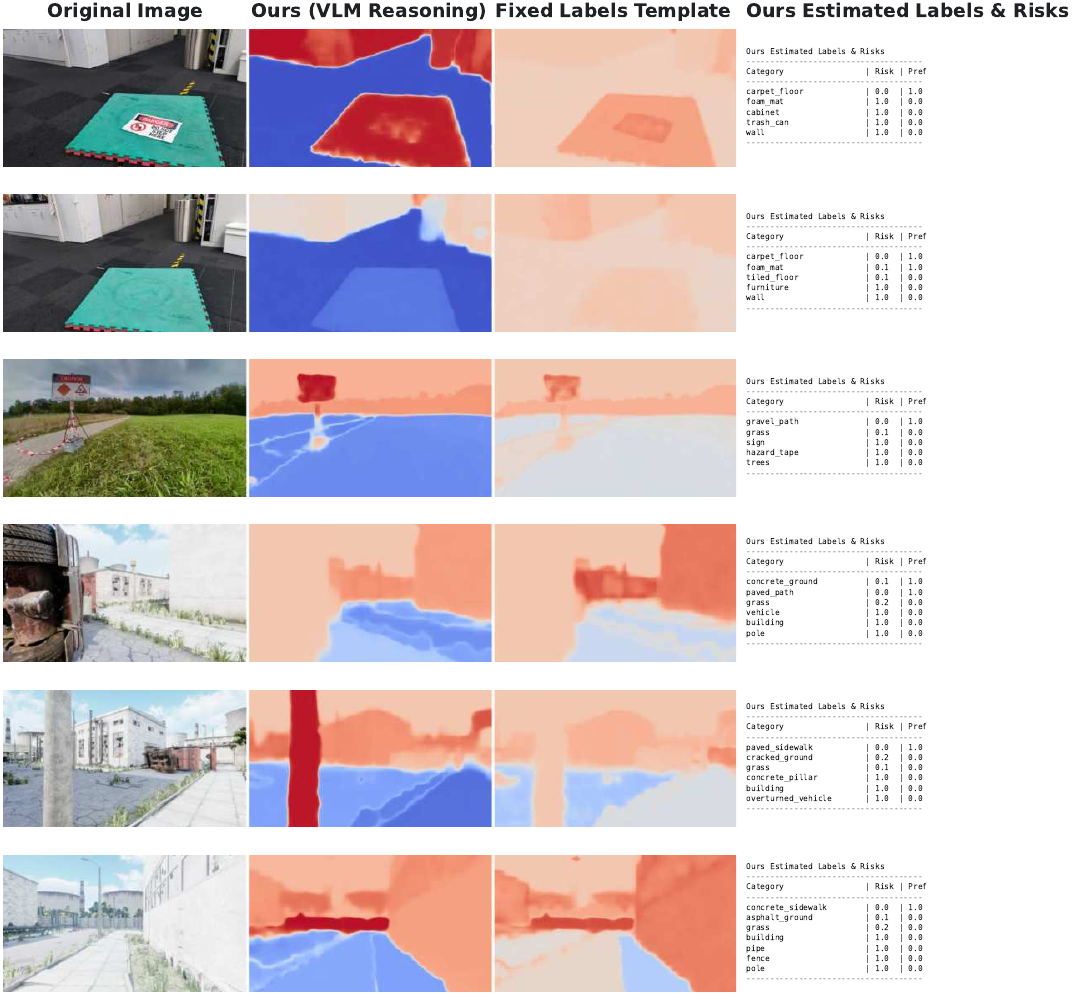}
  \caption{\textbf{Comparison with a fixed-label pipeline.} This comparison evaluates our dynamic, context-dependent scene-reasoning VLM framework against a Fixed Labels pipeline across evaluation datasets with a variety of indoor and outdoor scenes. 
  Each row shows the original image, the risk maps of our method and the Fixed Labels template, and the labels and risk values estimated by our method.}
  \label{fig:fixed_labels_comparison}
\end{figure*}

\begin{figure*}[htbp]
  \centering
  \includegraphics[width=\textwidth]{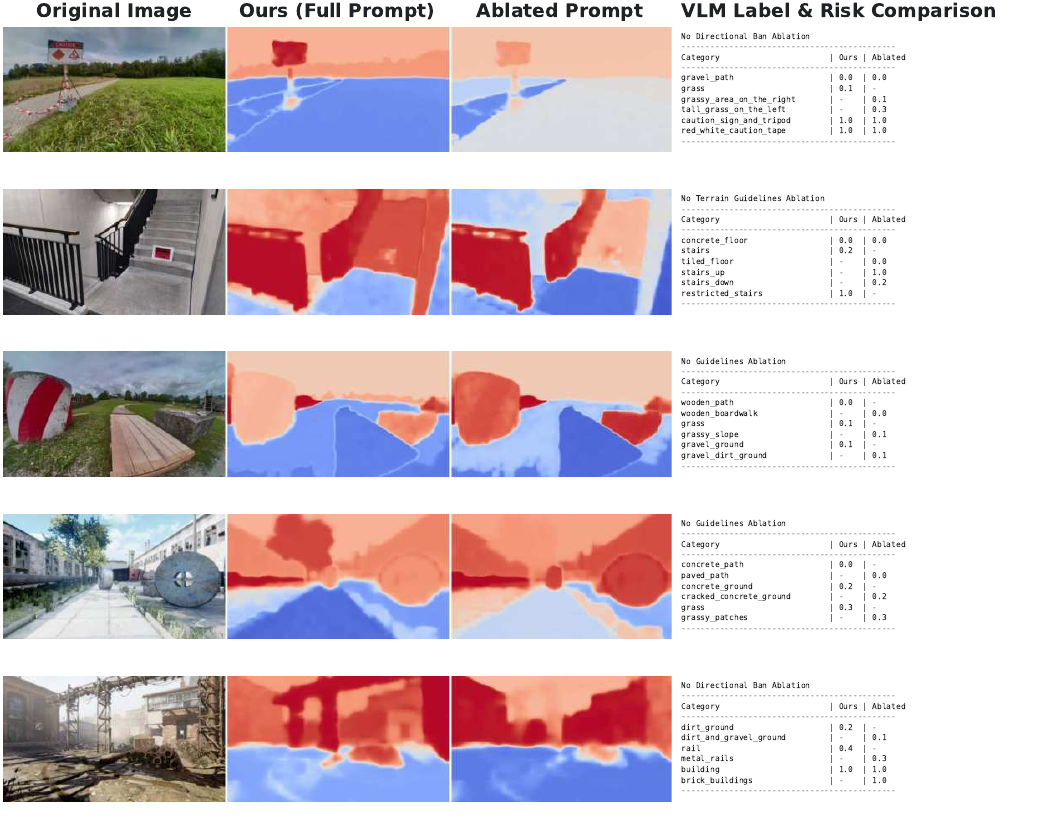}
  \caption{\textbf{Prompt ablation examples.} This prompt ablation study evaluates the visual and quantitative effect of omitting specific prompt constraints from the VLM prompt: the directional ban, the terrain guidelines, or all guidelines (see Supplementary Methods). The heading of each table names the ablation shown in that row. }
  \label{fig:prompt_ablation_study}
\end{figure*}

The result is shown in table~\ref{tab:prompt_ablation}.
It demonstrates that ROAM's open-vocabulary pipeline is robust to prompt variations. First, minor linguistic perturbations did not affect the performance much. The synonym-perturbed prompt maintained stable performance, indicating that the system is not overly sensitive to specific phrasing.
Second, ablating the structural constraints and guidelines degraded the performance, but not drastically (fig.~\ref{fig:prompt_ablation_study}). For example, completely removing the ``Directional Ban'' or ``Terrain Guidelines'' lowered the IoU slightly (dropping to 0.803 and 0.802, respectively, on the Sign dataset). This indicates that although our specific engineering guidelines help optimize the outputs for the downstream open-vocabulary segmentation model, the foundation model retains a resilient baseline of common-sense reasoning even without strict prompting rules.
Finally, the ablation confirms that using a ``Fixed Labels'' baseline did decrease the performance substantially (fig.~\ref{fig:fixed_labels_comparison}). Restricting the system to predefined categories caused the IoU to drop substantially to 0.665 on TartanGround and 0.612 on the Sign dataset. This quantitatively reinforces the core premise of ROAM: dynamic, open-vocabulary generation is a fundamental requirement for adapting to complex, unstructured real-world environments.

\begin{table*}[htbp]
\caption{\textbf{Prompt ablation and perturbation study.} Macro-averaged scores across the TartanGround (N = 372 images) and Sign (N = 12 images) benchmarks.}
\centering
\resizebox{0.75\textwidth}{!}{%
\begin{tabular}{l|ccc|ccc}
\hline
\multirow{2}{*}{Configuration} & \multicolumn{3}{c|}{TartanGround~\cite{patel2025tartanground}} & \multicolumn{3}{c}{Sign} \\
      & IoU & Prec. & Rec. & IoU & Prec. & Rec. \\
\hline
Baseline (Full Prompts)  & 0.827 & 0.840 & 0.985 & \textbf{0.904} & 0.924 & 0.977 \\
Perturbation (Synonyms)  & \textbf{0.842} & \textbf{0.853} & \textbf{0.987} & 0.867 & \textbf{0.925} & 0.936 \\
No Robot Description     & 0.819 & 0.829 & 0.988 & 0.878 & 0.906 & 0.952 \\
No Directional Ban       & 0.808 & 0.818 & 0.988 & 0.803 & 0.808 & \textbf{0.994} \\
No Terrain Guidelines    & 0.823 & 0.836 & 0.985 & 0.802 & 0.868 & 0.927 \\
No Guidelines            & 0.820 & 0.833 & 0.985 & 0.798 & 0.815 & 0.982 \\
Fixed Labels             & 0.665 & 0.668 & 0.996 & 0.612 & 0.612 & 1.000 \\
Fixed Labels (Simple)    & 0.587 & 0.587 & 0.999 & 0.699 & 0.736 & 0.955 \\
\hline
\end{tabular}
}
\label{tab:prompt_ablation}
\end{table*}

\subsubsection*{Robustness to nonexplicit linguistic cues in signage}
\label{sec:sup-ablation-explicit-text}

To quantitatively evaluate whether the system's performance degrades substantially in the absence of direct text instructions, we conducted an ablation study using the Sign dataset. For this study, we digitally altered the original images of the signs by removing all explicit navigational instructions (such as ``Do Not Walk on Path'' or ``Merge back to path''). The modified images retained only general ``DANGER'' or ``WARNING'' labels. We then processed these modified images through our \ac{VLM} pipeline.

\begin{figure*}[htbp] %
    \centering
    \includegraphics[width=1.0\linewidth]{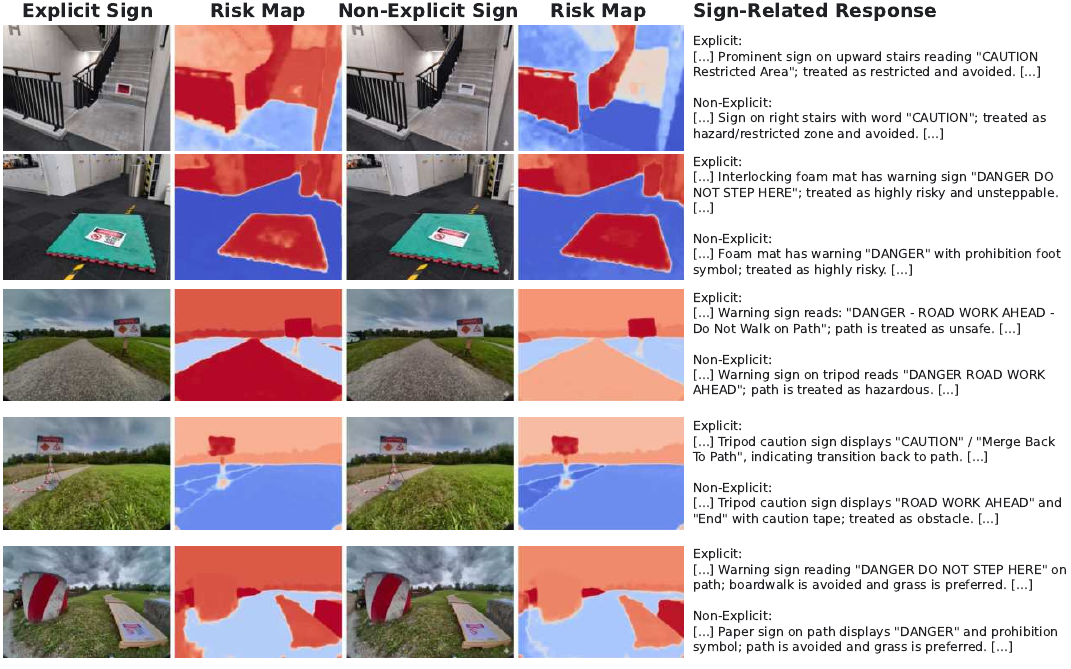}
    \caption{\textbf{Effect of the explicit instruction text within the sign images.} We removed the highly explicit text instructions (such as ``do not step here'' and ``danger mine'') from the sign regions (creating aligned non-explicit counterparts) and evaluated if the semantic risk prediction changed. }
    \label{fig:sign-explicit-comparison}
\end{figure*}

As shown in fig.~\ref{fig:sign-explicit-comparison} and detailed in table~\ref{tab:sign-comparison}, we found that the VLM still correctly grasps the intention of the signs without explicit navigational linguistic cues. It successfully correlated the general ``DANGER'' warning with the visual presence of environmental hazards, such as construction barriers and objects, thereby inferring the necessary navigational constraints without needing explicit directions. 

\begin{table*}[htbp]
\centering
\caption{\textbf{Performance comparison on explicit and nonexplicit sign datasets.} Macro-averaged scores across N = 12 paired sign scenes.}
\begin{tabular}{l|ccc|ccc}
\hline
\multirow{2}{*}{Configuration} & \multicolumn{3}{c|}{Explicit (SignDataset Baseline)} & \multicolumn{3}{c}{Non-Explicit (signs\_non\_explicit)} \\
      & IoU & Prec. & Rec. & IoU & Prec. & Rec. \\
\hline
Baseline (Full Prompts)  & \textbf{0.904} & 0.924 & \textbf{0.977} & 0.841 & \textbf{0.929} & 0.904 \\
\hline
\end{tabular}
\label{tab:sign-comparison}
\end{table*}

These results demonstrate that the system does not merely parse explicit text commands, but actively grounds general warnings into the physical scene using multimodal common-sense reasoning. Consequently, the navigation performance does not degrade substantially when encountering general warning signs, highlighting the robustness of the VLM's scene interpretation.

\subsection*{VLM stair traversability reasoning under explicit height constraints}
\label{sec:sup_stair_height}

ROAM does not rely on the VLM to judge centimeter-level physical
traversability; this is delegated to the 2.5D elevation map and the
learned navigation and locomotion policies. To make this design choice
concrete, we analyze how well current VLMs can enforce an explicit,
robot-specific height limit (such as ``stairs higher than 10\,cm cannot
be climbed'') from RGB images alone.

\subsubsection*{Dataset}
We collected 27 RGB images of staircases in diverse settings
(indoor and outdoor; varied lighting; concrete, carpeted, wooden, and
metal steps). The ground-truth riser heights were measured manually and
range from 13.0\,cm to 18.0\,cm. Each staircase was photographed
from multiple viewpoints and with different camera fields of view, so
that the same physical structure appears under varying perspective and
scale. Note that under a 10\,cm climbability limit every staircase in
the dataset is impassable, so the correct decision is coarse, whereas
under a 15\,cm limit the dataset contains climbable stairs
(13.0--14.5\,cm), boundary cases (15.5--16.0\,cm), and obviously
impassable stairs (17.0--18.0\,cm), requiring centimeter-level
discrimination.

\subsubsection*{Prompt variants}
We evaluated three model generations (Gemini~2.5, 3.0, and 3.5) under
three prompting conditions. We selected these models based on their availability at the time of the evaluation. The original condition uses the
standard ROAM prompt, which instructs the VLM to act as a navigation
planner for the quadrupedal robot and states that the robot can walk
over stairs, assigning them a low risk value depending on steepness; it
contains no height limit. The two constraint conditions append
an explicit rule to the robot description:

\begin{quote}
``You can walk over stairs or slopes, but they must be lower than
$\{10.0, 15.0\}$\,cm to be safe. If they are higher than
$\{10.0, 15.0\}$\,cm, they should be assigned a high risk value (1.0)
and unsteppable (0.0).''
\end{quote}

\subsubsection*{Newer VLMs follow coarse height limits but misclassify stairs near the limit}
\begin{figure*}[htbp]
  \centering
  \includegraphics[width=\linewidth,height=0.62\textheight,keepaspectratio]{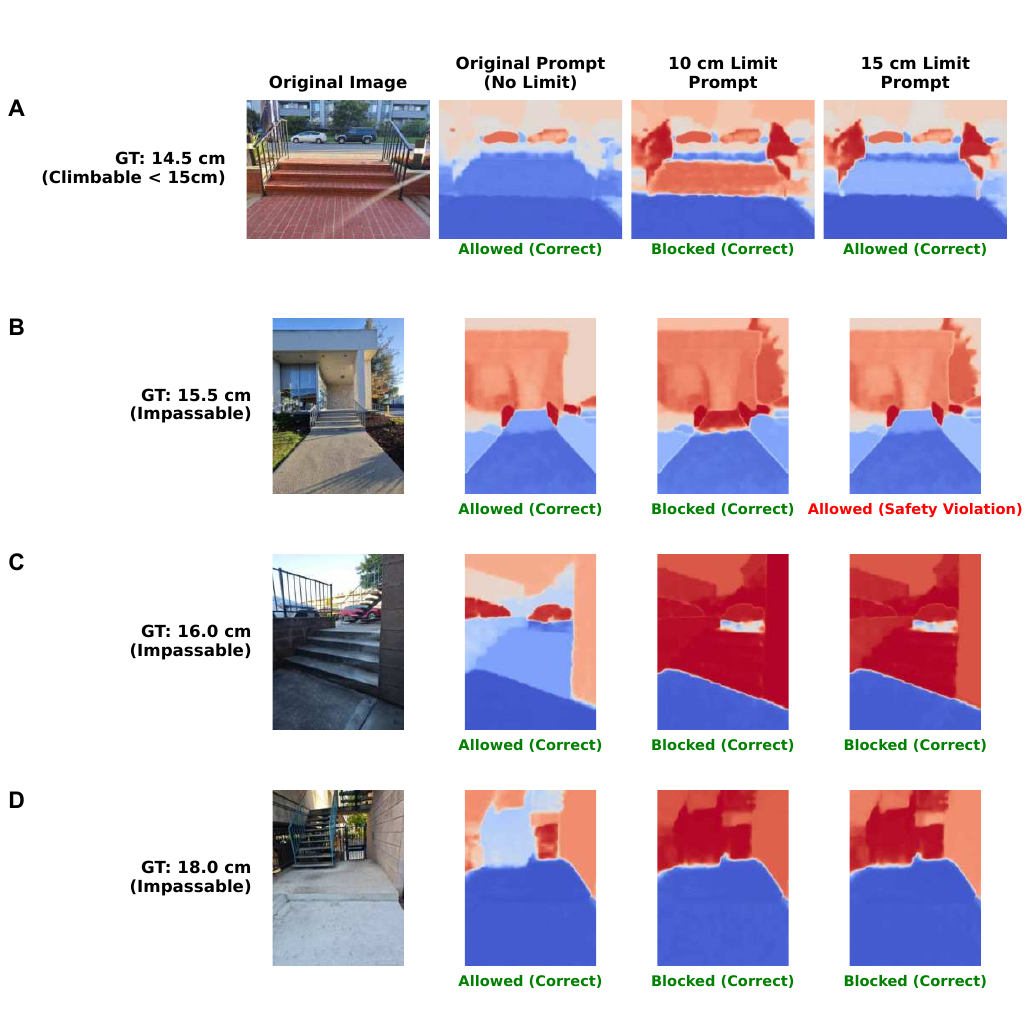}
  \caption{\textbf{Qualitative examples of VLM risk maps under explicit
  height constraints.} (A to D) Four staircases with measured riser heights
  of 14.5, 15.5, 16.0, and 18.0\,cm, each evaluated by Gemini~3.5 under three prompts: the original ROAM
  prompt (no height limit) and prompts with explicit 10\,cm and 15\,cm
  climbability limits. Blue indicates low assigned risk and red high
  risk. The text below each risk map states whether the staircase was allowed or blocked and whether this decision was correct.}
  \label{fig:sup_stair_examples}
\end{figure*}

\begin{figure*}[htbp]
  \centering
  \includegraphics[width=\linewidth]{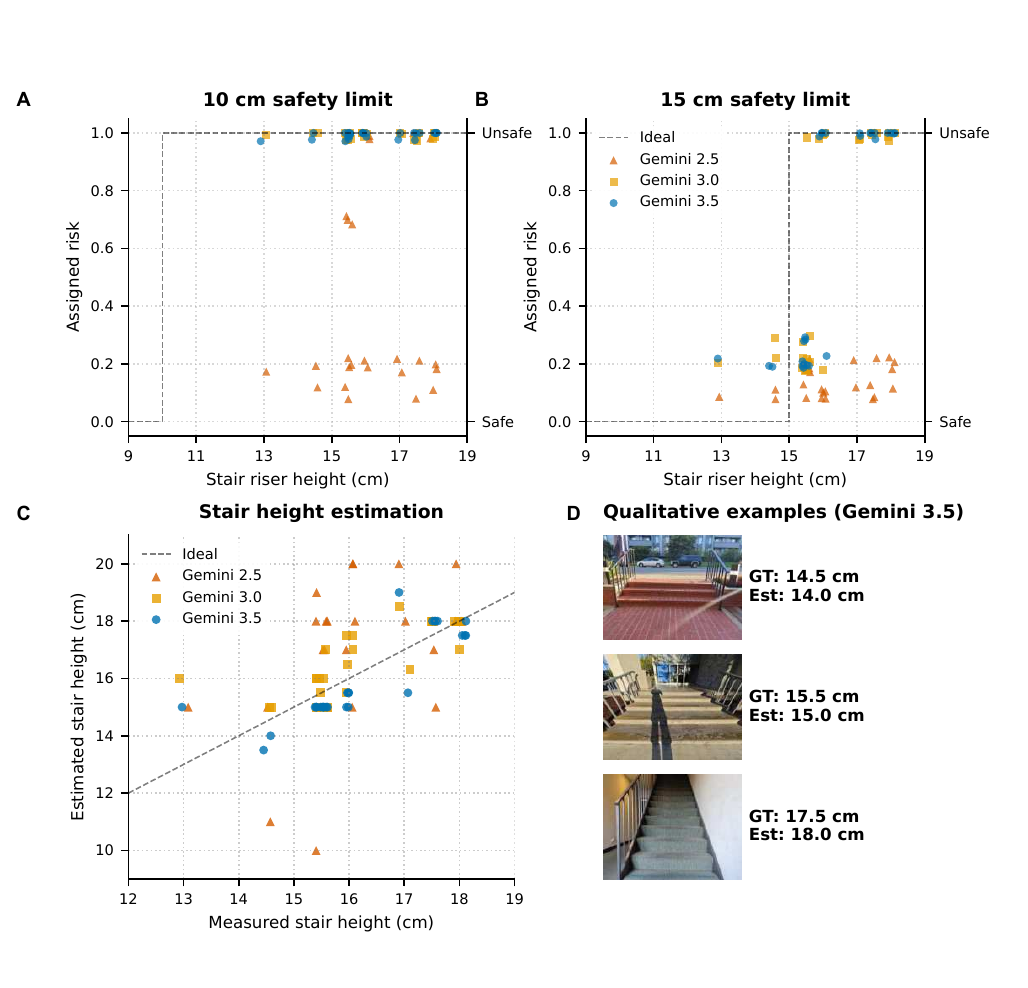}
  \caption{\textbf{Quantitative analysis of VLM stair-height
  reasoning.} (A and B) Risk assigned to staircases as a function of
  measured riser height under an explicit 10\,cm (A) and 15\,cm
  (B) safety limit, for three model generations (N = 27 images per model). (C) Directly estimated versus measured riser height without safety
  constraints. (D) Qualitative estimates from Gemini~3.5.}
  \label{fig:sup_stair_quant}
\end{figure*}

\paragraph*{Constraint adherence}
Fig.~\ref{fig:sup_stair_examples} shows qualitative risk maps and fig.~\ref{fig:sup_stair_quant}, A and B, the assigned risk as a function of measured riser height. Under the original prompt, all models correctly assigned low risk to the stairs, consistent with the prompt. Under the 10\,cm rule, the newer models (Gemini~3.0 and 3.5) correctly assigned the maximum risk of 1.0 and steppability of 0.0 to every staircase, whereas Gemini~2.5 frequently violated the explicit constraint, assigning risk values of 0.1--0.25 to stairs well above the limit. Under the 15\,cm rule, even the newest models misclassified boundary cases: staircases with 15.5--16.0\,cm risers were repeatedly assigned low risk, and in fig.~\ref{fig:sup_stair_examples}B a 15.5\,cm staircase is labeled traversable under the 15\,cm rule---a safety violation. Gemini~3.0 and 3.5 assigned high risk to all staircases well above the limit (17.0--18.0\,cm), whereas Gemini~2.5 assigned low risk to every staircase under the 15\,cm rule. 

\paragraph*{Direct stair-height estimation}
To isolate metric perception from decision making, we simply asked 
the models to estimate the riser height, without any other prompt.
Fig.~\ref{fig:sup_stair_quant}C plots them against the
ground truth. Gemini~2.5 produced estimates only weakly correlated with
the measured heights (mean absolute error $\sim$1.8\,cm, with outliers
exceeding 5\,cm), whereas Gemini~3.0 and 3.5 track the ground truth
substantially better ($\sim$0.7\,cm mean absolute error). Nevertheless,
errors of up to a few centimeters remain, and estimates for the same
staircase vary across viewpoints and camera fields of view. Notably,
even accurate-looking estimates (such as 15.0\,cm for a 15.5\,cm
staircase) can fall on the wrong side of a nearby safety threshold.

\paragraph*{Implications for ROAM}
First, the metric spatial understanding of VLMs is
improving rapidly across model generations: the newest models produce better-correlated height
estimates. Second, centimeter-level discrimination near a safety
threshold remains unreliable: residual estimation errors of a few
centimeters, their sensitivity to perspective and field of view, and
the tendency of estimates to drift toward a prompted threshold all
produce misclassifications precisely in the boundary regime. Delegating this decision to the VLM alone would produce
occasional safety violations. This supports the architecture of ROAM:
the VLM provides semantic risk priors, whereas the 2.5D elevation map and
the learned policies enforce the robot-specific physical limits. The
learned locomotion policy refuses untraversable stairs even when the
VLM fails to flag them, and when the VLM does assign high risk
correctly, the region is avoided earlier at the planning level.

\clearpage %

{

}

\clearpage
\onecolumn
\begin{longtable}{p{0.45\textwidth} c c c}
\caption{\textbf{Fixed Labels lookup table used as a baseline in the prompt ablation.} Risk, steppable, and preferred-area values assigned to each of the 55 predefined labels.}\label{tab:fixed_labels} \\
\toprule
\rowcolor{gray!30}
\textbf{Label} & \textbf{Risk} & \textbf{Steppable} & \textbf{Pref. Area} \\
\midrule
\endfirsthead

\multicolumn{4}{c}{{\bfseries \tablename\ \thetable{} -- Continued from previous page}} \\
\toprule
\rowcolor{gray!30}
\textbf{Label} & \textbf{Risk} & \textbf{Steppable} & \textbf{Pref. Area} \\
\midrule
\endhead

\midrule \multicolumn{4}{r}{{Continued on next page}} \\ \bottomrule
\endfoot

\bottomrule
\endlastfoot
\rowcolor{white} \multicolumn{4}{l}{\textbf{Solid, flat ground categories (Highly preferred, safe)}} \\
\hspace{1em} \texttt{concrete\_floor} & 0.0 & 1.0 & 1.0 \\
\hspace{1em} \texttt{concrete\_ground} & 0.0 & 1.0 & 1.0 \\
\hspace{1em} \texttt{paved\_path} & 0.0 & 1.0 & 1.0 \\
\hspace{1em} \texttt{wooden\_boardwalk} & 0.0 & 1.0 & 1.0 \\
\hspace{1em} \texttt{sidewalk} & 0.0 & 1.0 & 1.0 \\
\hspace{1em} \texttt{asphalt\_road} & 0.0 & 1.0 & 1.0 \\
\hspace{1em} \texttt{paved\_tile\_path} & 0.0 & 1.0 & 1.0 \\
\rowcolor{white} \multicolumn{4}{l}{\textbf{Safe, unpaved terrains (Partially preferred)}} \\
\hspace{1em} \texttt{dirt\_ground} & 0.1 & 1.0 & 0.8 \\
\hspace{1em} \texttt{hiking\_trail} & 0.1 & 1.0 & 0.8 \\
\hspace{1em} \texttt{trail\_ahead} & 0.1 & 1.0 & 0.8 \\
\hspace{1em} \texttt{soil} & 0.1 & 1.0 & 0.8 \\
\hspace{1em} \texttt{gravel\_ground} & 0.2 & 1.0 & 0.5 \\
\hspace{1em} \texttt{gravel\_path} & 0.2 & 1.0 & 0.5 \\
\hspace{1em} \texttt{cracked\_concrete\_ground} & 0.1 & 1.0 & 0.5 \\
\rowcolor{white} \multicolumn{4}{l}{\textbf{Stairs \& Slopes (Navigable with legged capabilities, but require caution)}} \\
\hspace{1em} \texttt{stairs} & 0.2 & 1.0 & 1.0 \\
\hspace{1em} \texttt{staircase} & 0.2 & 1.0 & 1.0 \\
\hspace{1em} \texttt{slope} & 0.2 & 1.0 & 0.8 \\
\hspace{1em} \texttt{ramp} & 0.2 & 1.0 & 0.8 \\
\hspace{1em} \texttt{curb} & 0.2 & 1.0 & 0.0 \\
\rowcolor{white} \multicolumn{4}{l}{\textbf{Low preferred / Soft/ Uneven terrains (Low priority)}} \\
\hspace{1em} \texttt{grass} & 0.3 & 1.0 & 0.0 \\
\hspace{1em} \texttt{grassy\_area} & 0.3 & 1.0 & 0.0 \\
\hspace{1em} \texttt{sand} & 0.4 & 1.0 & 0.0 \\
\hspace{1em} \texttt{sandy\_ground} & 0.4 & 1.0 & 0.0 \\
\hspace{1em} \texttt{rock} & 0.6 & 1.0 & 0.1 \\
\hspace{1em} \texttt{rocks} & 0.6 & 1.0 & 0.1 \\
\hspace{1em} \texttt{rocky\_ground} & 0.6 & 1.0 & 0.1 \\
\rowcolor{white} \multicolumn{4}{l}{\textbf{Hazard Categories (Always unsafe \& unsteppable)}} \\
\hspace{1em} \texttt{restricted\_stairs} & 1.0 & 0.0 & 0.0 \\
\hspace{1em} \texttt{wet\_cement} & 1.0 & 0.0 & 0.0 \\
\hspace{1em} \texttt{cardboard\_sign} & 1.0 & 0.0 & 0.0 \\
\hspace{1em} \texttt{cardboard} & 1.0 & 0.0 & 0.0 \\
\hspace{1em} \texttt{sign} & 1.0 & 0.0 & 0.0 \\
\hspace{1em} \texttt{wooden\_pallets} & 1.0 & 0.0 & 0.0 \\
\hspace{1em} \texttt{wooden\_pallet} & 1.0 & 0.0 & 0.0 \\
\hspace{1em} \texttt{water} & 1.0 & 0.0 & 0.0 \\
\hspace{1em} \texttt{river} & 1.0 & 0.0 & 0.0 \\
\hspace{1em} \texttt{puddle} & 1.0 & 0.0 & 0.0 \\
\rowcolor{white} \multicolumn{4}{l}{\textbf{Physical Obstacles / Structure (Always unsafe \& unsteppable)}} \\
\hspace{1em} \texttt{building} & 1.0 & 0.0 & 0.0 \\
\hspace{1em} \texttt{brick\_building} & 1.0 & 0.0 & 0.0 \\
\hspace{1em} \texttt{wall} & 1.0 & 0.0 & 0.0 \\
\hspace{1em} \texttt{concrete\_wall} & 1.0 & 0.0 & 0.0 \\
\hspace{1em} \texttt{fence} & 1.0 & 0.0 & 0.0 \\
\hspace{1em} \texttt{railing} & 1.0 & 0.0 & 0.0 \\
\hspace{1em} \texttt{concrete\_barrier} & 1.0 & 0.0 & 0.0 \\
\hspace{1em} \texttt{lamppost} & 1.0 & 0.0 & 0.0 \\
\hspace{1em} \texttt{pipes\_and\_valves} & 1.0 & 0.0 & 0.0 \\
\hspace{1em} \texttt{tree} & 1.0 & 0.0 & 0.0 \\
\hspace{1em} \texttt{vegetation} & 1.0 & 0.0 & 0.0 \\
\hspace{1em} \texttt{bush} & 0.8 & 0.0 & 0.0 \\
\hspace{1em} \texttt{dense\_vegetation} & 1.0 & 0.0 & 0.0 \\
\hspace{1em} \texttt{dense\_bushes} & 0.8 & 0.0 & 0.0 \\
\hspace{1em} \texttt{car} & 1.0 & 0.0 & 0.0 \\
\hspace{1em} \texttt{vehicle} & 1.0 & 0.0 & 0.0 \\
\hspace{1em} \texttt{pedestrian} & 1.0 & 0.0 & 0.0 \\
\hspace{1em} \texttt{person} & 1.0 & 0.0 & 0.0 \\
\hspace{1em} \texttt{other} & 1.0 & 0.0 & 0.0 \\
\end{longtable}

\begin{longtable}{p{0.45\textwidth} c c c}
\caption{\textbf{Fixed Labels (Simple) lookup table used as a baseline in the prompt ablation.} Risk, steppable, and preferred-area values assigned to each of the 9 predefined labels.}\label{tab:fixed_labels_simple} \\
\toprule
\rowcolor{gray!30}
\textbf{Label} & \textbf{Risk} & \textbf{Steppable} & \textbf{Pref. Area} \\
\midrule
\endfirsthead

\multicolumn{4}{c}{{\bfseries \tablename\ \thetable{} -- Continued from previous page}} \\
\toprule
\rowcolor{gray!30}
\textbf{Label} & \textbf{Risk} & \textbf{Steppable} & \textbf{Pref. Area} \\
\midrule
\endhead

\midrule \multicolumn{4}{r}{{Continued on next page}} \\ \bottomrule
\endfoot

\bottomrule
\endlastfoot
\rowcolor{white} \multicolumn{4}{l}{\textbf{Walkable ground}} \\
\hspace{1em} \texttt{ground} & 0.0 & 1.0 & 1.0 \\
\hspace{1em} \texttt{floor} & 0.0 & 1.0 & 1.0 \\
\rowcolor{white} \multicolumn{4}{l}{\textbf{Off-road paths / secondary ground}} \\
\hspace{1em} \texttt{trail} & 0.15 & 1.0 & 0.7 \\
\rowcolor{white} \multicolumn{4}{l}{\textbf{Walkable structured inclines / caution areas}} \\
\hspace{1em} \texttt{stairs} & 0.3 & 1.0 & 0.8 \\
\hspace{1em} \texttt{slope} & 0.2 & 1.0 & 0.8 \\
\rowcolor{white} \multicolumn{4}{l}{\textbf{Vegetation / foliage}} \\
\hspace{1em} \texttt{grass} & 0.4 & 1.0 & 0.2 \\
\rowcolor{white} \multicolumn{4}{l}{\textbf{Solid boundaries / building structures}} \\
\hspace{1em} \texttt{wall} & 1.0 & 0.0 & 0.0 \\
\rowcolor{white} \multicolumn{4}{l}{\textbf{General objects / boundaries / obstacles}} \\
\hspace{1em} \texttt{obstacle} & 1.0 & 0.0 & 0.0 \\
\rowcolor{white} \multicolumn{4}{l}{\textbf{Dynamic objects / people / vehicles}} \\
\hspace{1em} \texttt{person} & 1.0 & 0.0 & 0.0 \\
\end{longtable}

\clearpage
\paragraph{Data S1. Data underlying the figures and tables.}
Multi-sheet Excel file with the tabulated values underlying Figs.~\ref{fig:result_zermatt} and \ref{fig:vlm}C, Table~\ref{tab:vlm-comparison}, figs.~\ref{fig:zermatt-labels} to \ref{fig:mannequin} and \ref{fig:sup_stair_quant}, and tables~\ref{table:loc_quantitative} to \ref{tab:sign-comparison}. Each sheet is named after the figure panel or table it corresponds to, and P values are included where statistical tests were performed.

\end{document}